\documentclass[sigconf]{acmart}
\usepackage{wrapfig}
\usepackage{subcaption}
\usepackage{amsmath}
\usepackage{array}
\newcommand{\avsum}{\mathop{\mathpalette\avsuminner\relax}\displaylimits}
\makeatletter
\renewcommand\@formatdoi[1]{\ignorespaces}
\makeatother
\AtBeginDocument{%
  \providecommand\BibTeX{{%
    \normalfont B\kern-0.5em{\scshape i\kern-0.25em b}\kern-0.8em\TeX}}}
\newcolumntype{M}[1]{>{\centering\arraybackslash}m{#1}}
\usepackage{mathtools}

\makeatletter
\newcommand\avsuminner[2]{%
  {\sbox0{$\m@th#1\sum$}%
   \vphantom{\usebox0}%
   \ooalign{%
     \hidewidth
     \smash{\vrule height\dimexpr\ht0+1pt\relax depth\dimexpr\dp0+1pt\relax}%
     \hidewidth\cr
     $\m@th#1\sum$\cr
   }%
  }%
}
\makeatother

\setcopyright{none}
\renewcommand\footnotetextcopyrightpermission[1]{} 
\acmConference[]{}{}{}
\begin{document}

\title{Efficient Human-in-the-loop System for Guiding DNNs Attention}

\author{Yi He}
  \email{he-yi@jaist.ac.jp}
\affiliation{%
  \institution{Japan Advanced Institute of Science and Technology}
  \city{Ishikawa}
  \country{Japan}
}

\author{Xi Yang}
  \email{earthyangxi@gmail.com}
\authornote{Corresponding author. Part of this work was done at The University of Tokyo.}
\affiliation{%
  \institution{Jilin University}
  \city{Jilin}
  \country{China}}

\author{Chia-Ming Chang}
  \email{info@chiamingchang.com}
\affiliation{%
  \institution{The University of Tokyo}
  \city{Tokyo}
  \country{Japan}
}

\author{Haoran Xie}
  \email{xie@jaist.ac.jp}
\affiliation{%
 \institution{Japan Advanced Institute of Science and Technology}
 \city{Ishikawa}
 \country{Japan}}

\author{Takeo Igarashi}
\email{takeo@acm.org}
\affiliation{%
  \institution{The University of Tokyo}
  \city{Tokyo}
  \country{Japan}}
\renewcommand{\shortauthors}{He and Yang, et al.}

\begin{abstract}
Attention guidance is used to address dataset bias in deep learning, where the model relies on incorrect features to make decisions. Focusing on image classification tasks, we propose an efficient human-in-the-loop system to interactively direct the attention of classifiers to regions specified by users, thereby reducing the effect of co-occurrence bias and improving the transferability and interpretability of a deep neural network (DNN). Previous approaches for attention guidance require the preparation of pixel-level annotations and are not designed as interactive systems. We herein present a new interactive method that allows users to annotate images via simple clicks. Additionally, we identify a novel active learning strategy that can significantly reduce the number of annotations. We conduct both numerical evaluations and a user study to evaluate the proposed system using multiple datasets.
Compared with the existing non-active-learning approach, which typically relies on considerable amounts of polygon-based segmentation masks to fine-tune or train the DNNs,
our system can obtain fine-tuned networks on biased datasets in a more time- and cost-efficient manner and offers a more user-friendly experience.
Our experimental results show that the proposed system is efficient, reasonable, and reliable. Our code is publicly available at \url{https://github.com/ultratykis/Guiding-DNNs-Attention}.
\end{abstract}





\begin{teaserfigure}
  \centering
  \includegraphics[width=\textwidth]{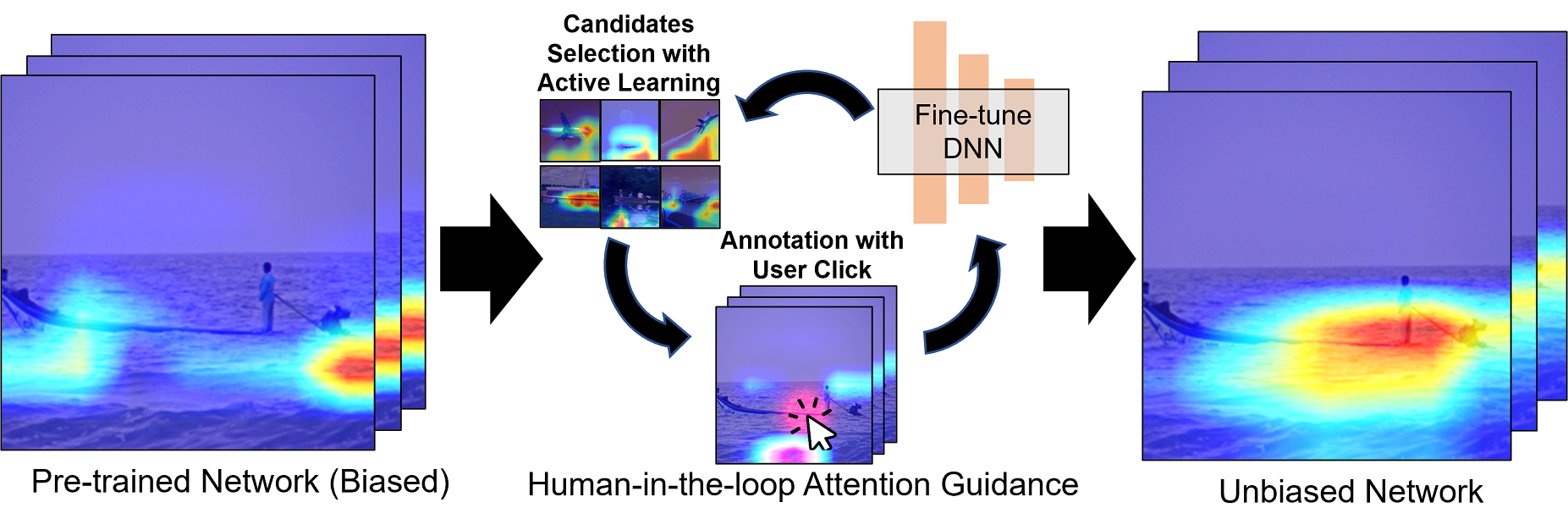}
  \caption{Overview of our system. By designing a single-click attention-directing user interface and an attention-based active learning strategy, our system significantly reduces the number of annotation operations and images used to direct the attention of deep neural network classifiers.}
  \label{fig:teaser}
\end{teaserfigure}

\settopmatter{printfolios=true}
\maketitle
\pagestyle{plain}
\section{Introduction}
\label{sec:intro}

The transferability and interpretability of deep neural networks (DNNs) are critical for various deep-learning applications. Correct and clear evidence allows users to be more confident of their network decisions, and renders the pretrained networks easier to deploy across different datasets. However, the datasets we collect often have various biases. One such bias, i.e., co-occurrence bias, is due to insufficient negative examples in a dataset~\cite{torralba2011unbiased}. The strong correlation among several feature elements results in features that are to be extracted, which are often accompanied by other features in the same context. Based on the characteristics of a DNN, we should not expect a network trained using datasets with significant biases to make decisions based only on the necessary features. This has been proven in many experiments~\cite{torralba2011unbiased, Zhang2018ExaminingCR, zhao2019large}, and we use feature visualization techniques~\cite{guidotti2018survey} to show the attention of DNNs, which indicates the features used to make decisions.

\begin{figure}[h]
\centering
  \includegraphics[width=0.8\linewidth]{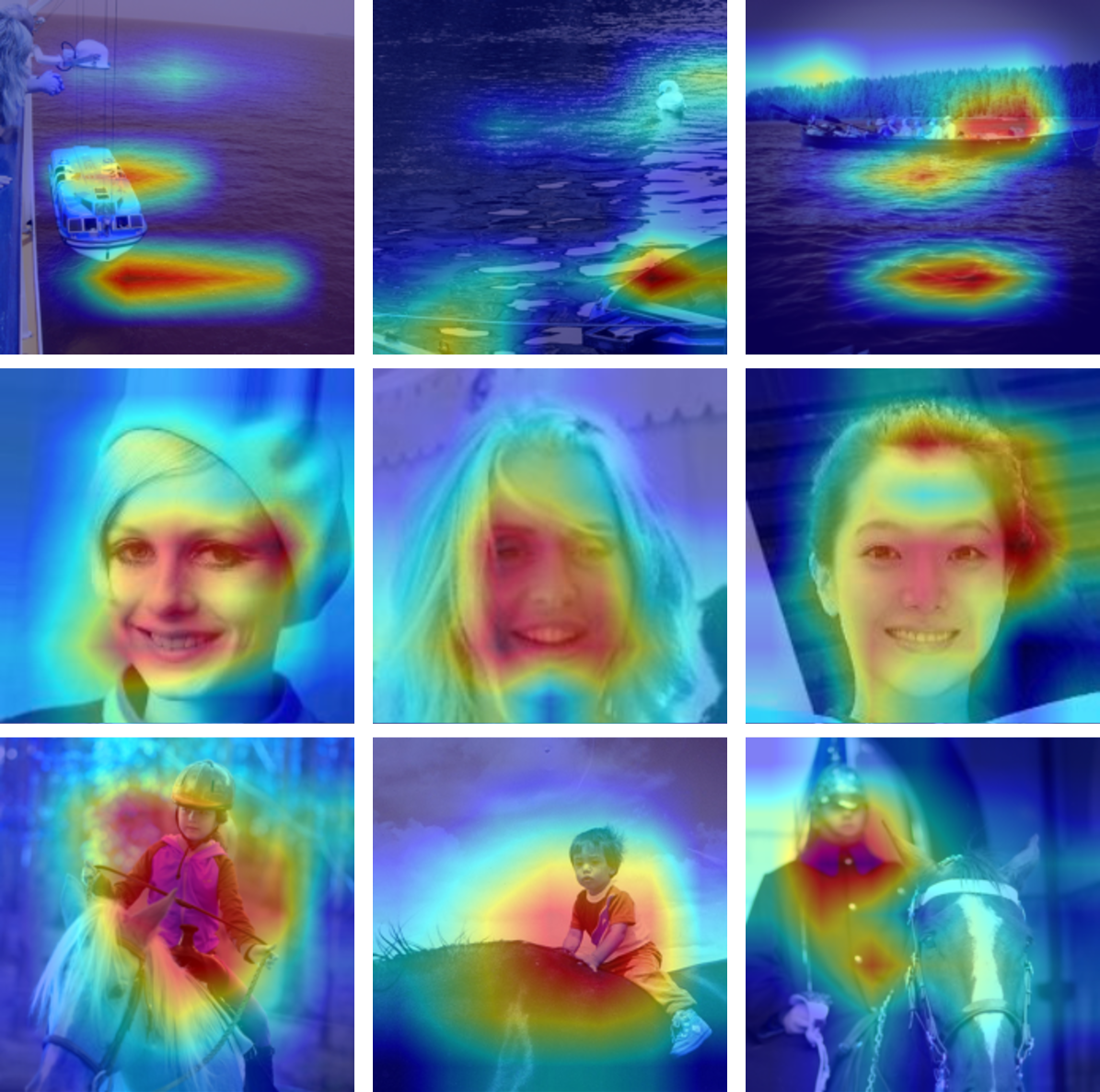}
  \caption{Examples showing co-occurrence bias in multiple datasets. Importance of regions from high to low for prediction is indicated from red to blue. Images from COCO dataset are shown in the first row. Results of feature visualization (Grad-CAM) show that the pre-trained DNN classifies ``boat'' by focusing on water waves, the contour of the coast, or the border between boat and water, instead of the boat itself. Images from the CelebA dataset are shown in the second row. Here, the pre-trained network-classified ``lipstick'' attribute focuses not only on the region of the mouth, but also on the eye and eyebrow. Finally, images from the AwA2 dataset are shown in the third row. In this case, the pre-trained network classifies `horse' by relying on both a horse and a human.}
\label{fig:biased}
\end{figure}

\paragraph{Examples.} 
Figure~\ref{fig:biased}shows examples of the co-occurrence bias problem. \textit{Boat-water problem}~\cite{torralba2011unbiased}. In the COCO dataset~\cite{lin2015microsoft}, all boats were captured in a river or sea, and the networks recognized water waves or coastal contours instead of various boat shapes to classify boat images. \textit{Lipstick problem}~\cite{Zhang2018ExaminingCR}. In the Large-scale CelebFaces Attributes (CelebA) dataset \cite{liu2015faceattributes}, the attributes ``Wearing Lipstick'' and ``Heavy Makeup'' typically occur simultaneously with high probability. Most people in the images, who not only apply lipstick, but also makeup on other facial regions, will only be labeled by the ``Wearing Lipstick'' attribute. Thus, the network recognizes ``Wearing Lipstick'' by relying on the makeup applied on several regions of the face, such as the eyes, eyebrows, and mouth. \textit{Horse-human problem}~\cite{zhao2019large}. In the AwA2 dataset~\cite{8413121}, humans typically appear in the images of horses; thus, to classify the horse class, the networks rely not only on the features of the horses, but also on those of humans.  

Biased representations are not easily eliminated from a pretrained network because the interpretability of current DNN networks is inferior. The features are extracted in high dimensions and extremely difficult to disentangle. For such intra-class bias, mining valid information is difficult using a simple approach of randomly selecting images from different categories during training. To eliminate these biases, a direct solution is to modify the training dataset to balance them~\cite{khosla2012undoing,selvaraju2017grad,zhao2019large}. However, this method requires a significant number of manual operations for each bias, and there are issues in the following two solutions: First is to remove images that have biases, which will significantly reduce the dataset size and result in the loss of many features. Second is to increase the number of negative samples. For the lipstick problem, for example, to balance the effects of other parts on the network, samples with makeup on other facial sections without lipstick must be added. However, negative samples are difficult to obtain, and the presence of numerous negative samples may prevent the network from extracting features. Furthermore, the increase in training data necessitates a more complex network model, more effective devices, and a longer training time.

Another approach is to focus the system (attention) on a specific area of the images. Manual and automatic methods have been developed. In manual approaches, for example, more weights are manually assigned to the important region; however, determining the size and shape of the region is difficult and may result in overfitting. Fully automatic approaches, such as the attention mechanism~\cite{fukui2018attention} or self-attention~\cite{li2018tell}, cannot easily eliminate biased representations from a pretrained network because the extracted features cannot be disentangled easily, and self-guidance by a soft mask is ineffective if the region of interest (ROI) overlaps with the attention map. Bahng et al.~\cite{bahng2020learning} reduced a network's dependency on the bias representation by encouraging it to be different from networks with small receptive fields that can easily capture texture features. However, disentangling object-level bias is difficult. In this case, supervised methods~\cite{li2018tell, yang2019directing} force the attention of the network on the regions provided by additional
information. However, they rely on pixel-level annotations, which incur high human costs. This additional information (attention positions on images) is difficult to obtain unless images with appropriate annotations are acquired in advance~\cite{aggarwal2020active} or if the elements can be detected using existing methods~\cite{yang2019directing}.

Therefore, we herein propose an interactive method to improve the efficiency of supervised methods to guide the attention of a pretrained network for resolving co-occurrence bias. It comprises the following two elements: The first is a clicking interface for region selection. The user specifies the regions where the classifiers should or should not focus on using simple clicks (users are not obligated to specify all ROIs), and the system fine-tunes the classifier such that it focuses on or avoids the specified regions. To achieve this simple operation, we designed an adjusted loss function by combining the region and center differences based on Gaussian mixture models (GMMs). The second is a new active learning strategy that substantially reduces the number of annotation images required to obtain the expected results. The main contributions of this study are as follows:

\begin{itemize}
    \item We propose a human-in-the-loop system to efficiently reduce the effect of dataset co-occurrence bias on pre-trained DNNs, thereby improving the transferability and interpretability of the latter.
    
    \item We design a new user interface to allow users to guide the attention of a network by simply clicking the ROIs and a new loss function such that the network undergoes fine-tuning. We investigate a new active learning strategy based on the distributions of calculated feature maps.
    
    \item We conduct a user study to compare existing methods as baselines to demonstrate the efficiency of our proposed system. This includes a comparison of the proposed click-based annotation approach with a classical polygon-based approach, as well as a comparison of attention directing using the designed active learning mechanism and a non-active learning method.
\end{itemize}

\section{Related studies}

In addition to the studies mentioned in the Introduction, we extensively surveyed studies pertaining to human-computer interactions, machine learning, and computer vision.

\subsection{Human-in-the-loop Frameworks}

Human-in-the-loop is an effective approach for connecting human knowledge and machine-learning models~\cite{amershi2014power}. Several applications of this approach have been reported in the literature. 
Nakano et al.~\cite{nakano2020interactive} proposed a framework, known as interactive deep singing-voice separation to allow users to fine-tune a deep-learning model and improve the separation accuracy of an input song.
Shimizu et al.~\cite{shimizu2020design} designed a framework that allows users to create and edit design adjectives to explore high-dimensional spaces.
Silva et al.~\cite{silva2021encoding} presented a reinforcement learning technique for the intelligent initialization of neural network weights and architectures by encoding domain knowledge.
Hilgrad et al.~\cite{hilgard2021learning} introduced a human decision function into a learning framework to optimize its representation.
Mishra et al.~\cite{mishra2021designing} introduced an interactive tool to support non-experts in transfer learning.
Wang et al.~\cite{wang2021autods} introduced an automated machine learning system to support data science projects.
In addition, several interaction techniques~\cite{benenson2019large, agustsson2019interactive, majumder2019content} have been investigated to improve the performance of deep learning models for image segmentation through human annotations.
Liu et al.~\cite{liu2019deep} proposed a deep reinforcement active learning system to maximize person re-identification performance.
Various interactive applications~\cite{koyama2017sequential, koyama2020sequential, chiu2020human} have been designed for computational design based on human-in-the-loop optimization methods.

\subsection{Dataset Bias}

Dataset bias was introduced by Torralba and Efros~\cite{torralba2011unbiased}, who analyzed and summarized various dataset biases by comparing and evaluating a set of well-established datasets. 
Additionally, Tommasi and Tuytelaars~\cite{tommasi2014testbed} performed a test to analyze a cross-dataset comprising 12 image datasets. 
Researchers have attempted to detect or eliminate biases learned by networks.
Stock et al.~\cite{stock2018convnets} proposed the use of adversarial examples to uncover undesirable biases learned by a model.
Kim et al.~\cite{kim2019learning} proposed a regularization term based on mutual information to unlearn the target bias from training data.
Maksai and Fua~\cite{maksai2019eliminating} exploited appearance-based features to eliminate exposure bias in multiple-object tracking tasks.
Tian et al.~\cite{tian2018eliminating} proposed a person-region guided pooling DNN based on human parsing maps to eliminate background bias in person re-identification.
Hendricks et al.~\cite{hendricks2018women} introduced two complementary loss terms, i.e., appearance confusion loss and confidence loss, to enable a network to learn less gender bias. 
Li and Vasconcelos~\cite{li2019repair} introduced a dataset resampling procedure based on the formulation of bias minimization as an optimization problem.

\subsection{Feature Visualization}

The biased representations of a pretrained DNN are caused by dataset biases. Feature visualization methods allow one to verify whether pretrained DNNs have biased representations~\cite{adebayo2018sanity}.
The simplest method is to achieve this is visualize the weights of the feature maps directly, as introduced by Krizhevsky et al.~\cite{krizhevsky2012imagenet}.
Simonyan et al.~\cite{Simonyan2013DeepIC} generated an artificial image, which is representative of a class of interest, and computed an image-specific class saliency map to highlight the areas of interest.
Zeiler and Fergus~\cite{zeiler2014visualizing} visualized the activities of each layer by applying deconvolution.
Springenberg et al.~\cite{springenberg2015striving} proposed guided backpropagation to visualize features learned through DNNs.
Yosinski et al.~\cite{DBLP:journals/corr/YosinskiCNFL15} introduced an excellent software tool by combining several different visualization methods.
Zhou et al.~\cite{zhou2016learning} generated class activation maps (CAMs) using global average pooling in DNNs to highlight discriminative object sections detected by the DNN.
Selvaraju et al.~\cite{selvaraju2017grad} proposed a gradient-weighted class activation mapping (Grad-CAM) technique that realizes CAM applications without modifying the model architecture.

\begin{figure*}[t]
\centering
  \includegraphics[width=\linewidth]{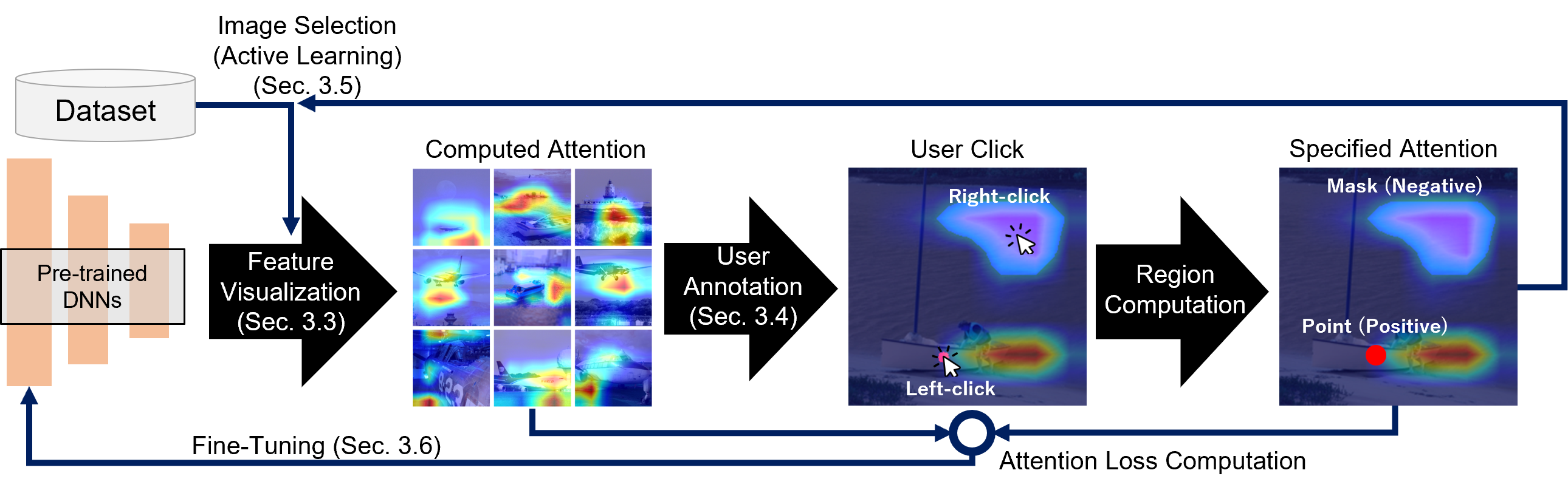}
  \caption{Overview of proposed method. For the same input image, the fine-tuned unbiased network can output a non-biased attention map compared with the input.}
\label{fig:overview}
\end{figure*}

\subsection{Active Learning}

Active learning aims to reduce the number of annotations by selectively labeling the most informative samples. It comprises three main settings: membership query synthesis, stream-based selective sampling, and pool-based active learning~\cite{settles2009active}.
Pool-based active learning has been widely applied, three main sampling methods exist: uncertainty, diversity, and query-by-committee-based methods. In this study, we focus on pool-based active learning methods for image-related tasks to introduce state-of-the-art studies. 
For example,
Sener and Savarese~\cite{sener2017active} defined the active learning problem as core set selection and investigated a core set problem for DNNs.
Sinha et al.~\cite{sinha2019variational} used Variational Autoencoders to learn a low-dimensional latent space using labeled and unlabeled data.
Yoo et al.~\cite{yoo2019learning} proposed a loss prediction module that is directly applicable to any task involving DNNs to predict the target losses of unlabeled inputs.
Zhang et al.~\cite{zhang2020state} proposed a state relabeling adversarial active learning model that considers both annotation and state information to derive the most informative unlabeled samples.
Agarwal et al.~\cite{agarwal2020contextual} introduced contextual diversity to unify model prediction uncertainty with diversity based on spatial and semantic contexts.
Siddiqui et al.~\cite{siddiqui2020viewal} exploited the viewpoint consistency in multiview datasets to propose an active learning strategy for semantic segmentation.
Aggarwal et al.~\cite{aggarwal2020active} adapted active learning for imbalanced visual tasks by modifying acquisition functions.
Gudovskiy et al.~\cite{gudovskiy2020deep} proposed a low-complexity method for feature density matching using a self-surprised Fisher kernel and pseudo-label estimators for biased datasets.
Yuan et al.~\cite{yuan2021multiple} proposed a multiple-instance active object detection approach for selecting informative images from an unlabeled set.

\subsection{Interpreting DNNs}

For DNNs to be interpretable, they must not only make a decision, but also explain the reason for such a decision~\cite{arrieta2020explainable}. Insufficient interpretability renders it difficult for DNN decisions to be fully trusted on many critical tasks, although they have achieved remarkable performance.
Both statistical methods~\cite{koh2017understanding, zhang2018visual} and visualization tools~\cite{LIU201748} have been studied to open the black box to improve the network interpretability. 
Besides, Ribeiro et al.~\cite{ribeiro2016should} proposed local interpretable model-agnostic explanations, which attempt to fit a surrogate function that approximates a neural network. The description of the surrogate function may be used to explain the output decisions of any model.
Visualizing attention at regions where the extracted features are located is an intuitive method that allows users to understand the evidence for making a decision pertaining to image-related tasks.

\subsection{Attention Guidance}

Biases can be eliminated using automatic approaches, as introduced in Section~\ref{sec:intro}, among which those presented in \cite{fukui2018attention, li2018tell} are representative examples. Pixel-level supervision~\cite{li2018tell} is a typically used approach for addressing biased attention in DNN models used in the machine-learning field. This approach allows the DNN models to focus on the desired regions; however, it relies on the ground truth of image segmentation, which is obtained by drawing a polygon box along the boundary of each object on an image with significant human resources. The attention mechanism-based method~\cite{fukui2018attention}, which introduces an attention branch network that extracts the effective weight by generating an attention map for a visual explanation based on response-based visual explanation in a supervised learning manner, can similarly reduce network bias. However, as they rely on extracted image features that are typically difficult to disentangle, biased representations cannot be effectively eliminated. Yang et al.~\cite{yang2019directing} used landmarks as ground truth to guide the attention of pre-trained models; however, this method is limited to scenarios in which landmarks can be detected using existing tools such as face recognition. Shen et al.~\cite{shen2021human} claimed that they implemented an interactive framework to obtain human feedback for attentional guidance. However, they neither performed an extensive evaluation nor a user study, which renders the efficiency of the framework questionable. In addition, the feature visualization method used requires changing the architecture of the pretrained network.

\begin{figure*}[t]
\begin{center}
  \includegraphics[width=0.9\linewidth]{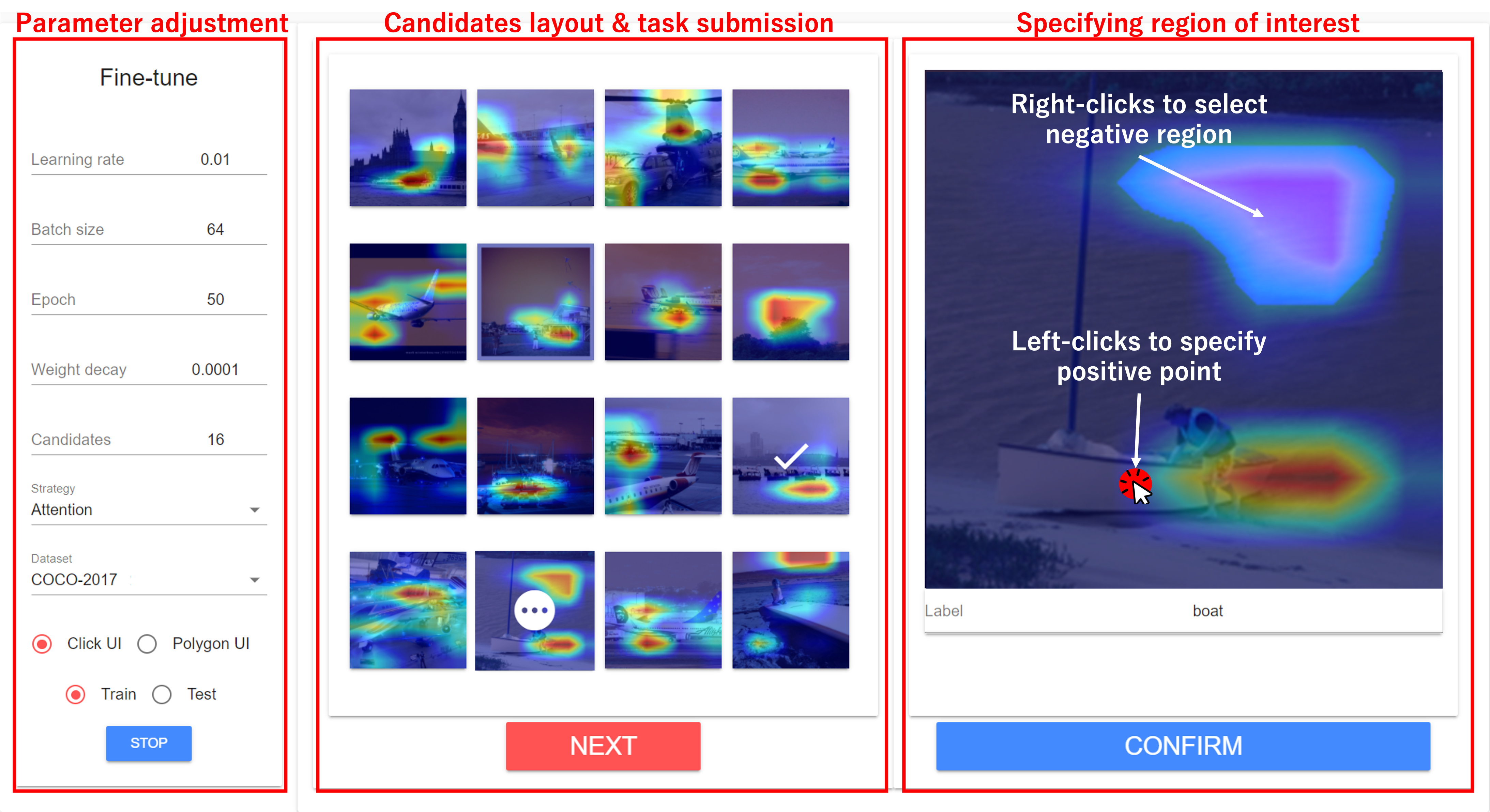}
\end{center}
   \caption{The click-based user interface is developed by combining an active learning interactive mechanism. The user can modify the parameters, select a strategy to be used in active learning, and select a dataset for fine-tuning. When the user clicks the START button (it changes to the STOP button after starting), the back-end system returns the selected images and waits for the user to confirm the annotation and apply fine-tuning. The user can refine the attention map in the workspace on the right side of the user interface by clicking the left mouse button to specify centers of the positive attention region, the right mouse button to decide the negative region, and the middle mouse button to cancel all operations on this image.}
\label{fig:ui}
\end{figure*}

\section{System design}

\subsection{Design Rationale}

Deep-learning networks make decisions completely by relying on prior information. Because feature visualization allows humans to understand the basis of network reasoning, human knowledge must be embedded to disentangle co-occurring features from datasets. Whereas a human-in-the-loop system can efficiently allow users to interactively and targetedly rectify pretrained networks, human knowledge is an expensive resource if dense annotations are involved. Therefore, we investigated active learning and user interface design to significantly reduce the cost of labor. The active learning strategy identifies commonalities in incorrect samples to prevent the user from providing redundant annotations, and the efficient user interface allows the user to annotate rapidly and accurately.

\subsection{Overview}

As shown in Figure~\ref{fig:overview}, for a pretrained DNN model with biased representations, our system selects informative images from a specified dataset using the active learning strategy and visualizes the attention regions. Subsequently, the user annotates the selected images by clicking on the positive or negative attention regions as user-specified attention. Next, the difference between the predicted and user-specified attention is calculated as a new loss function term to fine-tune the pretrained network and allow it to focus on the correct regions. Our loss function only requires the user to click within the positive/negative area and not the exact pixels. Thus, biased representations are efficiently eliminated from the network.

\subsection{Attention Visualization}

Based on comparing the results of several feature visualization approaches, we decided to employ Grad-CAM~\cite{selvaraju2017grad} in our study. Grad-CAM uses class-specific gradient information to localize important regions. For an image and a specified class, the image is passed through the network to achieve activation. Subsequently, a one-hot vector marked by the class is backpropagated to compute the gradients entering the convolutional layer. The average of these gradients is computed to obtain the weights of each feature map. The heat map showing the detected region is a weighted combination of the feature maps.

\subsection{User Interface Design}
\label{subsec:uidesign}

To improve annotation efficiency and reduce the workload of the annotator, we propose a custom-designed user interface that can support user interactions via the active learning mechanism, as shown in Figure~\ref{fig:ui}. For wider applicability, the user interface was designed as a web application based on Vuejs~\cite{you2020vuejs}. The back-end system was implemented using Flask~\cite{flask}.

\textbf{Parameter adjustment.} 
We provide a parameter panel on the left for the user to adjust the dataset, data selection strategies, training parameters, and number of candidates that should be labeled at a time. In addition, the user can switch to the training or test mode at any time to assess the performance of the current fine-tuned model.

\textbf{Candidate layout and task submission.} 
We prepared a layout that displays task submissions in the middle. The user can manage any candidate image as desired and submit the annotation results. The submitted images are denoted by a checkmark to allow users to confirm their annotation progress.

\textbf{Specifying ROI.} 
We propose a single-click-type operation for specifying the ROIs. The left user clicks the centers of the positive attention regions to specify the region wherein the network should focus. If the user thinks that the network is focused on the incorrect region, then the user right-clicks on the incorrect attention regions to remove the effect on these regions such that the network can make a decision.

When users complete image labeling and click the CONFIRM button, the workspace is emptied and the images being processed in the middle area of the candidate images are marked as completed, thus allowing the users to easily verify their work progress. The remaining candidate images are displayed by clicking the NEXT button. After all candidate images are labeled, the NEXT button changes to the FINE-TUNE button. Clicking on the FINE-TUNE button advances the process to the next round model fine-tuning. This process continues until the user confirms that a more satisfactory result has been achieved.

\begin{figure*}[t]
    \centering
    \includegraphics[width=\linewidth]{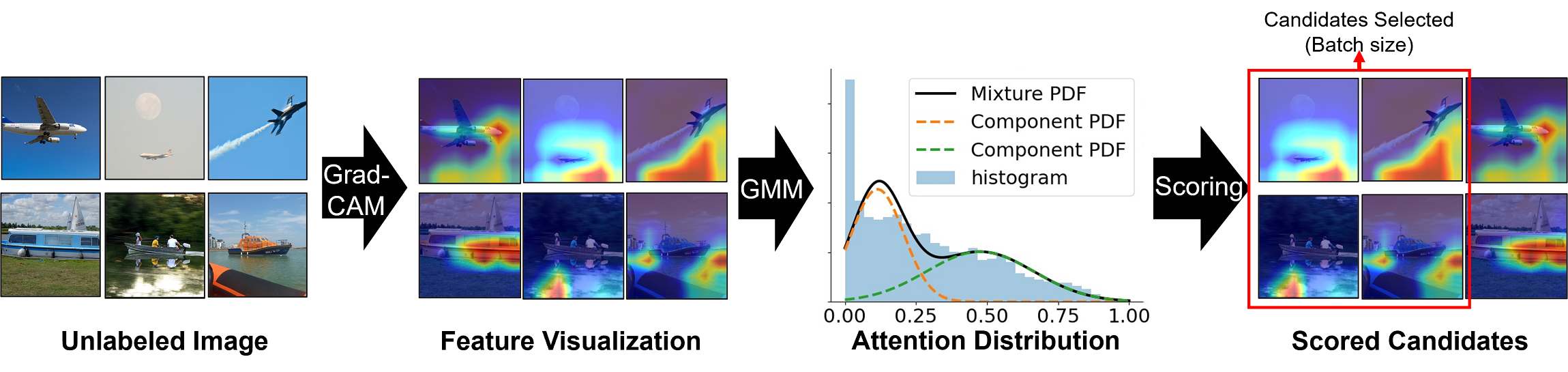}
    \caption{Proposed attention based data selection strategy designed for attention directing.}
    \label{fig:strategy_gmm}
\end{figure*}

\subsection{Active Learning Strategy}

Active learning is used in several applications of machine learning. It uses the minimum number of labeled samples to maximize the performance gains of the model during each training epoch. During this process, the strategy used to sample the data affects the training time required to obtain the desired results. The main strategies typically used in deep active learning include random selection, entropy~\cite{6889457}, and the K-center greedy~\cite{sener2017active} algorithm.

For our task, we designed a strategy based on the attention distribution of unlabeled data to select informative images, as shown in Figure~\ref{fig:strategy_gmm}. We believe that the features of the most worthwhile annotated images are similar to the common characteristics of dataset images. Thus, we used a GMM~\cite{murphy2012machine} to determine the most important attention maps for user direction. A GMM is a probabilistic model that assumes that all data points are sampled from a combination of a finite number $K$ of components. In this model, each component (with a component weight $\pi_k$) is a multivariate Gaussian distribution with mean $\mu_k$ and covariance matrix $\sum_k$. The model exhibits the following form:

\begin{equation}
\label{equ:gmm}
    p(x_i|\theta)=\sum^K_{k=1}\pi_k\mathcal{N}(x_i|\mu_k,\textstyle\sum_k)
\end{equation}

We introduced the GMM to our system to score unlabeled images based on the diversity of co-occurrence and then corrected the activation value distributions. In our implementation, the $K$ was set to two, which represents positive and negative features, and each component had its own diagonal covariance matrix $\sum_k$. We used the expectation-maximization (EM) algorithm~\cite{dempster1977maximum} to train our model and the number of EM iterations was set to $100$. The predicted attention maps of the unlabeled images were fitted to the GMM as input to obtain a trained model. Subsequently, we calculated the log-likelihood of each sample $x$ from $score_{x}$ using Equation~(\ref{equ:strategy}). For simplicity, we consolidated all GMM parameters into $\theta$. $score_{x}$ is the probability of the observed sample under the fitted $g_{\theta}$, which can be sorted in the descending order. Subsequently, we selected the top batch size number of images to be annotated by the user.

\begin{equation}
    \label{equ:strategy}
    \textup{score}_{x} = \textup{ln}(g_{\theta}(x))
\end{equation}

\subsection{Fine-tuning Approach}
\label{subsec:ft}

\begin{figure}[t]
    \centering
    \includegraphics[width=0.9\linewidth]{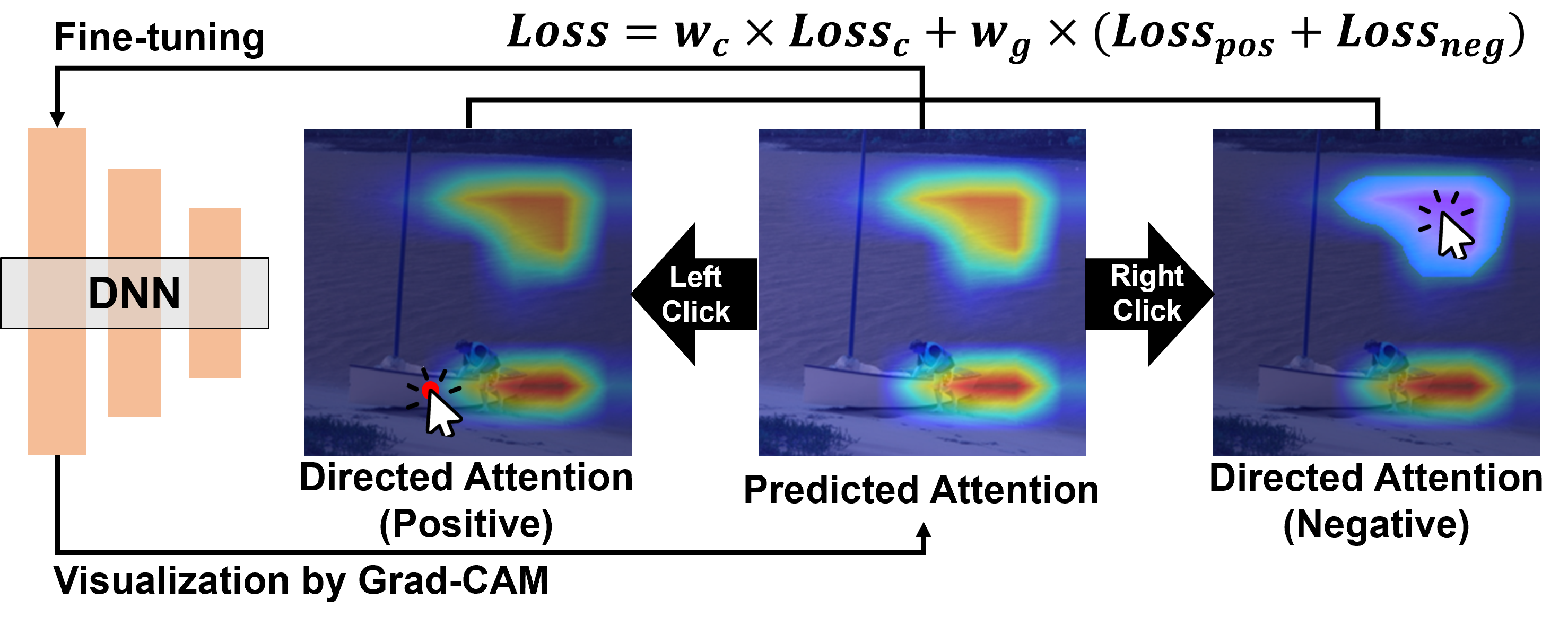}
    \caption{Computation of fine-tuning loss.}
    \label{fig:attention_loss}
\end{figure}

The pretrained network was fine-tuned using the loss function, which is a combination of weighted guidance loss $loss_g$ and classification loss $loss_c$, computed as \(Loss = w_g*loss_g + w_c*loss_c\) (see Figure~\ref{fig:attention_loss}). Here, the classification loss $loss_c$ is calculated as pretraining using the binary cross entropy between the predicted scores and labels. The guidance loss $loss_g$ is computed as $loss_g = loss_{pos} + loss_{neg}$, where the positive loss $loss_{pos}$ and negative loss $loss_{neg}$ are directed by the user interaction. Parameters $w_g$ and $w_c$ are the weights to balance $loss_g$ and $loss_c$, respectively. 

Because our system allows users to specify multiple positive points, 
we obtained $loss_{pos}$ by calculating the difference between the center of the user-directed positive points ($pp$) and the barycenter of the predicted attention map ($am$), as shown in Equation~(\ref{equ:positive}).
Simultaneously, we incorporated a super-pixel method~\cite{felzenszwalb2004efficient} to segment the predicted attention map to allow the user to select negative attention regions ($nr$) via simple clicks. 
Here, $Coord$ refers to the coordinates from the attention map and the value of the activation map is denoted as $Val$.
Because we expect the activation of the negative attention region to be approximately zero, we calculated the sum of the values of the attention map located in the directed negative regions as $loss_{neg}$, as expressed in Equation~(\ref{equ:negative}).

Notably, the users were not required to click on all ROIs; additionally, the user can specify arbitrary regions or remain idle.

\begin{equation}
    \label{equ:positive}
    \textup{loss}_{\textup{pos}}  = \left\| \frac{\sum\limits_{i\in \textup{am}} (\textup{Coord}_i \cdot \textup{Val}_i)}{\sum\limits_{i\in \textup{am}} \textup{Val}_i} \ - \ \avsum_{j\in{\textup{pp}}} \textup{Coord}_j \right\| 
\end{equation}

\begin{equation}
    \label{equ:negative}
    \textup{loss}_{\textup{neg}} = \sum_{k\in \textup{nr}} \textup{Val}_k
\end{equation}

\section{Implementation}

\subsection{Dataset}

We evaluated our method using three typically used datasets that are affected by co-occurrence bias: COCO-2017~\cite{lin2015microsoft}, CelebA~\cite{liu2015faceattributes}, and AwA2~\cite{8413121}. We classified the datasets into three sets: training, test, and validation. We use the ``airplane'' and ``boat'' categories from COCO-2017; ``Wearing Lipstick'' from CelebA; ``humpback whale'', and ``horse'' from AwA2. Table~\ref{tab:datasets} lists the compositions of the datasets used.

\begin{table}[h]
    \centering
    \caption{Composition of various datasets used in our implementation.}
     \begin{tabular}{| c | c | c | >{\centering}m{11mm} | >{\centering}m{13mm} | c |} 
     \hline
      & \multicolumn{2}{|c|}{COCO-2017} & CelebA & \multicolumn{2}{|c|}{AwA2} \\ [0.5ex] 
     \hline
     Set & airplane & boat & Wearing Lipstick & humpback whale & horse \\
     \hline\hline
     Train & 2388 & 2420 & 13436 & 651 & 1316 \\ 
     Test & 598 & 605 & 3839 & 150 & 264 \\ 
     Validation & 97 & 121 & 1920 & 50 & 65\\ [1ex] 
     \hline
     \end{tabular}
     \label{tab:datasets}
\end{table}

\subsection{Training}

The proposed system was implemented on a Linux system computer with two NVIDIA RTX 3090 graphics cards, an Intel® Xeon(R) W-2223 CPU, and 64GB memory. We used ResNet-18~\cite{he2016deep} (provided by PyTorch~\cite{NEURIPS2019_9015}), which contained 18 convolutional layers, to perform classification task. We pretrained ResNet-18 on our three prepared datasets at a learning rate of 0.1 and a weight decay of 0.0005. We used stochastic gradient descent as the optimizer in the pretraining process and applied early stopping to obtain the pretrained model. The pretraining epoch was set to 150, the batch size was 128. In the fine-tuning process, we used a learning rate of 0.01 to fine-tune the pretrained model. The values of $w_g$ and $w_c$ for the fine-tuning loss were set to $0.75$ and $0.25$, respectively.
The batch size was 128, and the epoch was set to 150.

\section{Quantitative Evaluation of Active Learning Strategy}
We conducted a numerical evaluation by manually guiding the attention of a biased pre-trained network using our system to compare our active learning (candidate selection) method against random selection and two existing selection methods, including entropy~\cite{NIPS2013_b6f0479a, 10.5555/1613715.1613855}, and diversity-based algorithms. For the entropy-based selection strategy, we calculated the entropy-based on the softmax output vector values. For the diversity-based approach, we used K-Center-Greedy~\cite{sener2018active}, which is a core-set algorithm for data sampling. 

We evaluated the effectiveness of our proposed method in terms of the fine-tuned accuracy and visualization results. The same pre-trained network was fine-tuned using different strategies on the categories of ``boat'' and ``airplane'' categories within the same limited amount of time (240 min), as shown in Table~\ref{tab:acc_comp_coco}. The proposed method achieved the highest accuracy. 

Additionally, we conducted tests on the ``Wearing Lipstick'' and ``Heavy Makeup'' categories, as shown in Table~\ref{tab:acc_comp_celeba}. To demonstrate the performance of each fine-tuned network on images based on only the ``Wearing Lipstick'' attribute, we developed a new test set by removing images with the ``Heavy Makeup'' attribute in the previous test set. The results show that the network used in the proposed method learned more exact features and achieved better transferability than other three selection strategies.

\begin{table}
    \centering
     \makeatletter\def\@captype{table}\makeatother \caption{The mean accuracies of 3 trials on COCO-2017 achieved by each strategy.}
     \label{tab:acc_comp_coco}
     \begin{tabular}{c | M{2cm} | M{2cm}}
     \hline
     Strategy & airplane & boat \\ [0.5ex] 
     \hline\hline
     Pre-trained & 88.155\% & 80.331\% \\ 
     Random & 88.365\% & 82.975\% \\ 
     Entropy & 88.574\% & 84.023\% \\ 
     Diversity & 88.679\% & 84.682\% \\ 
     Attention (ours) & \textbf{88.994\%} & \textbf{84.792\%} \\ [1ex] 
     \hline
    \end{tabular}
\end{table}

\begin{table}
    \centering
        \makeatletter\def\@captype{table}\makeatother \caption{The mean accuracies of 3 trials on CelebA achieved by each strategy.}
     \label{tab:acc_comp_celeba}
     \begin{tabular}{c | M{2cm} | M{2cm}} 
     \hline
     Strategy & Wearing Lipstick & Wearing Lipstick only\\  [0.5ex] 
     \hline\hline
     Pre-trained & 82.204\% &  73.038\%\\ 
     Random & 82.998\% & 73.084\% \\ 
     Entropy & 83.274\% &  73.176\% \\ 
     Diversity & 83.614\% &  73.245\% \\ 
     Attention (ours) & \textbf{83.730\%} &  \textbf{73.359\%} \\ [1ex] 
     \hline
      \end{tabular}
\end{table}

\begin{figure*}[h]
\centering
\begin{subfigure}{0.49\linewidth}
  \centering
  \includegraphics[width=\textwidth]{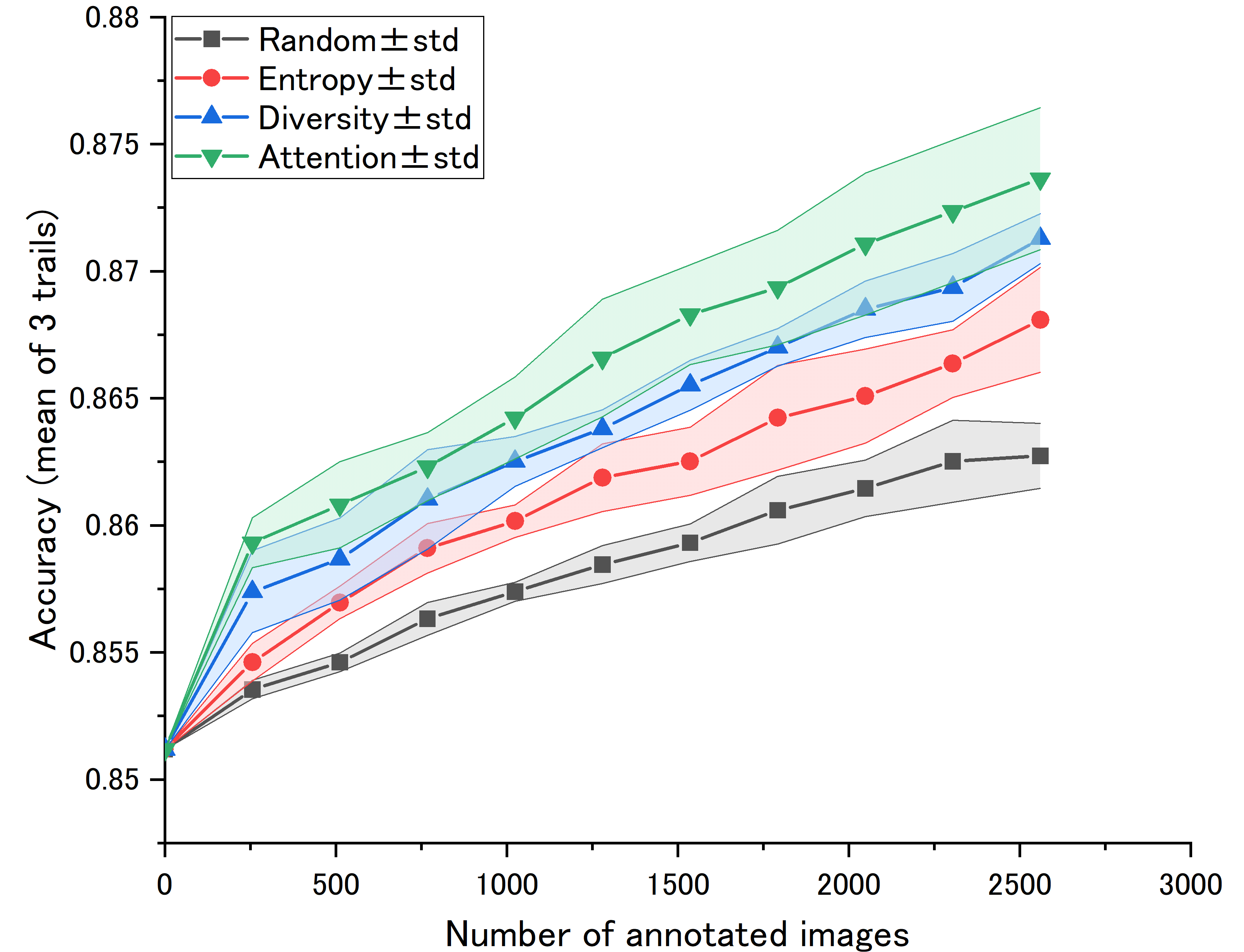}
    \caption{COCO-2017}
    \label{fig:accuracy_curve_1}
\end{subfigure}
\begin{subfigure}{0.49\linewidth}
  \centering
    \includegraphics[width=\textwidth]{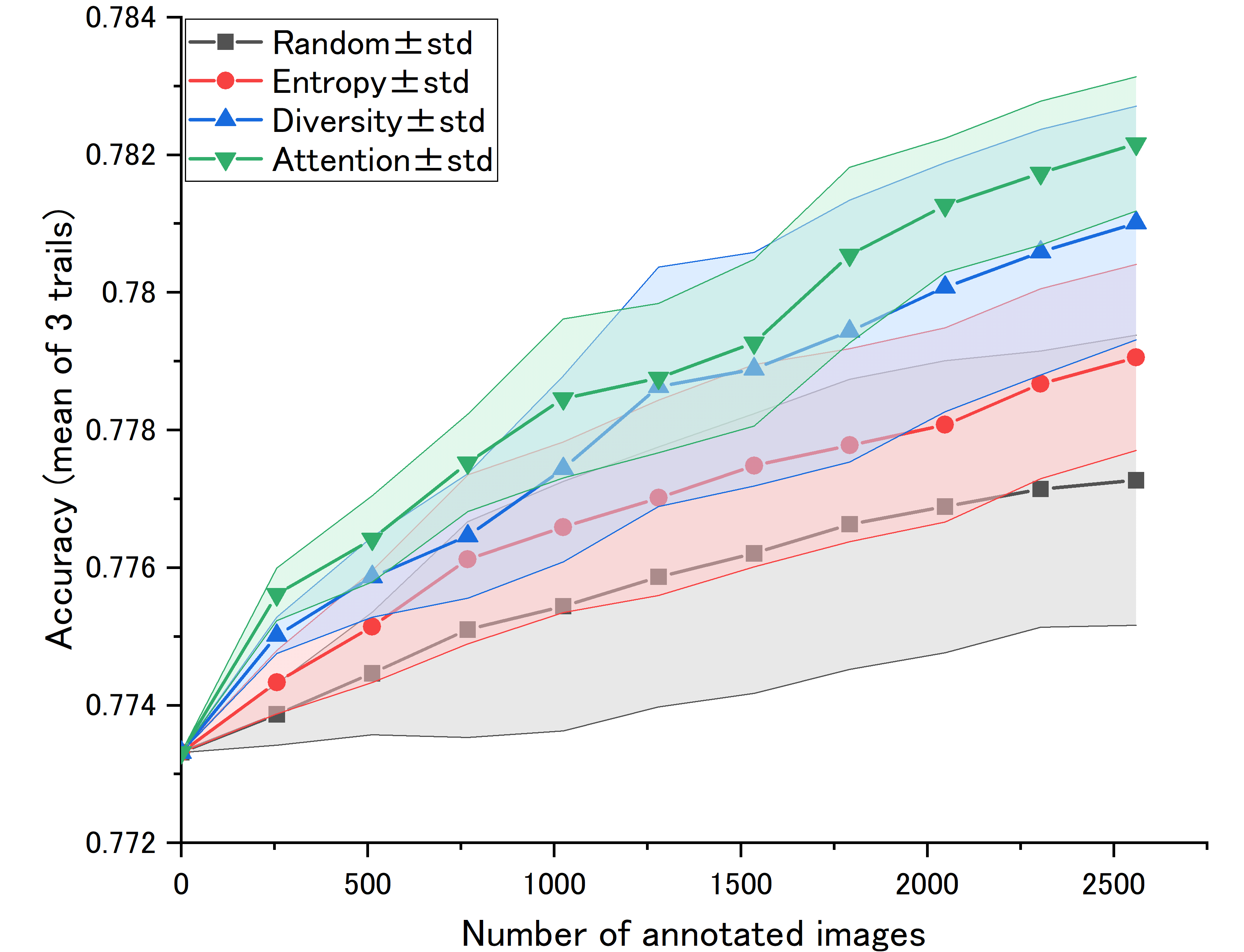}
    \caption{CelebA}
    \label{fig:accuracy_curve_2}
\end{subfigure}
\caption{Accuracy variations of fine-tuning progress. (a) Results on COCO-2017; (b) Results on CelebA. For the active learning results obtained using different strategies when applying the COCO-2017 and CelebA datasets, we labeled 2560 images during 10 cycles. The mean accuracy and the standard deviation of three trials are shown in the figure.}
\label{fig:accuracy_curve}
\vspace{-1.5em}
\end{figure*}

To demonstrate the effect of our proposed method on active learning under the same annotation workload, we visualize and show the active learning curve for the different strategies used in Figures~\ref{fig:accuracy_curve}(a) and \ref{fig:accuracy_curve}(b). As shown in the figures, both the entropy- and diversity-based approaches can achieve better results faster than the random sampling approach on different datasets. During the final cycle of active learning, the random-based approach achieved total accuracies of 86.273\% and 77.727\%, whereas the entropy- and diversity-based approaches achieved accuracies of 86.808\% and 87.128\% on COCO-2017, 77.905\% and 78.101\% on CelebA, respectively. By contrast, the accuracies of our proposed attention-based maneuver were 87.363\% and 78.216\% on COCO-2017 and CelebA, respectively.

\begin{figure}[htp]
    \centering
    \begin{subfigure}{\linewidth}
    \centering
    \includegraphics[width=\linewidth]{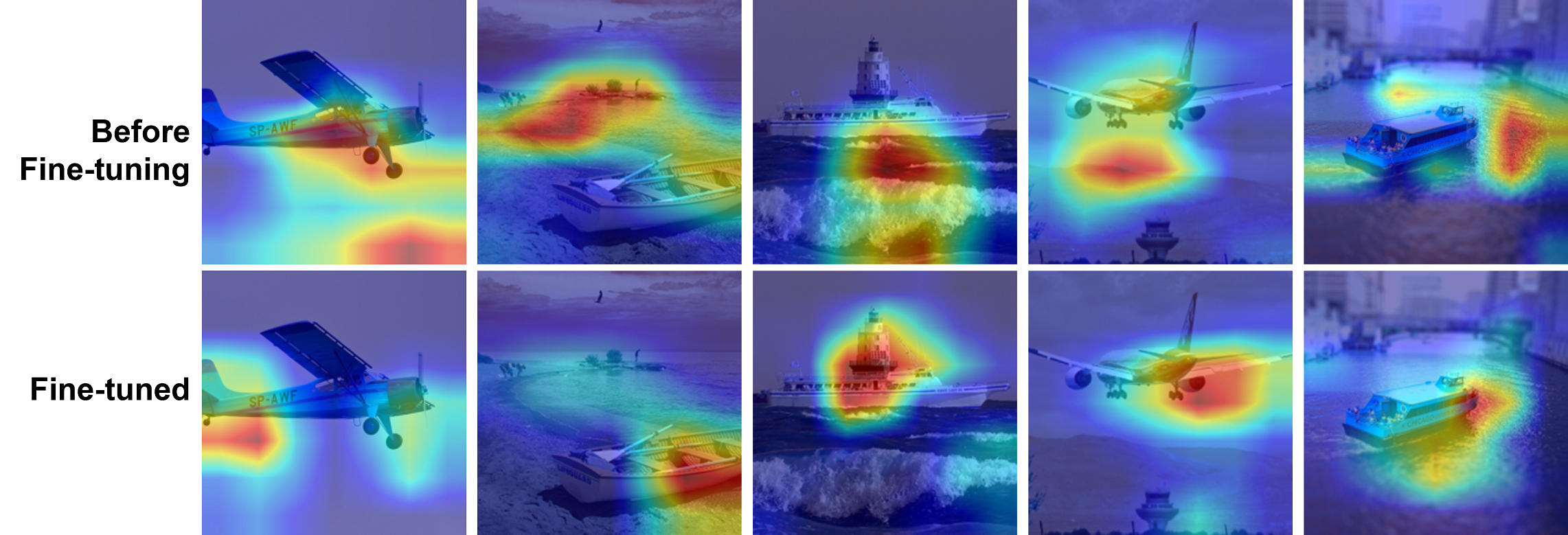}
    \caption{``boat'' and ``airplane'' from COCO-2017}
    \label{fig:result_examples_coco}
    \end{subfigure}
    \begin{subfigure}{\linewidth}
    \centering
    \includegraphics[width=\linewidth]{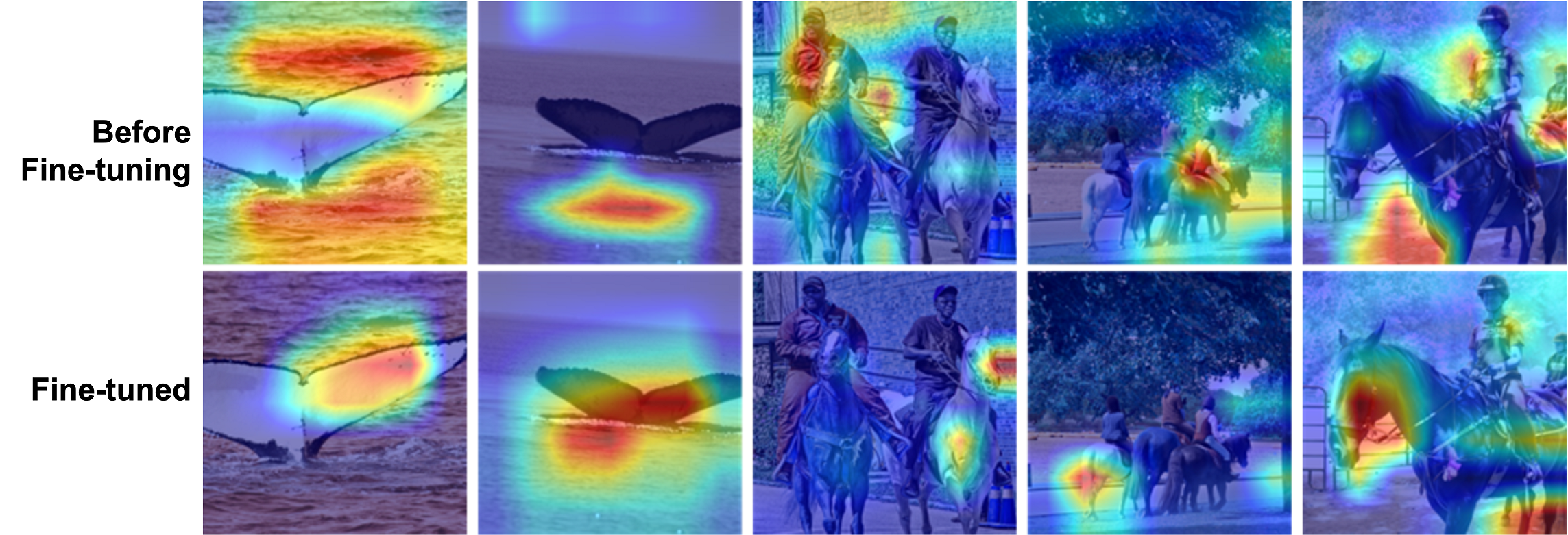}
    \caption{``humpback whale'' and ``horse'' from AwA2}
    \label{fig:result_examples_awa2}
    \end{subfigure}
    \begin{subfigure}{\linewidth}
    \centering
    \includegraphics[width=\linewidth]{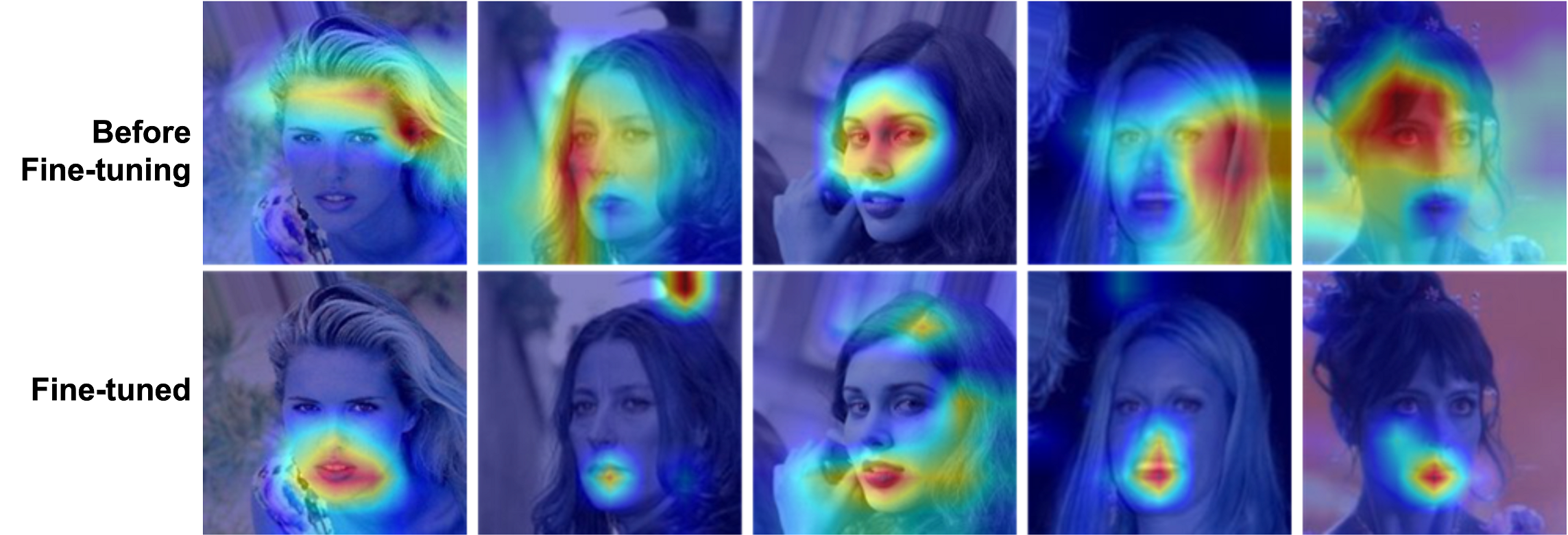}
    \caption{``Wearing Lipstick'' from CelebA}
    \label{fig:result_examples_celeba}
    \end{subfigure}
    \begin{subfigure}{\linewidth}
    \centering
    \includegraphics[width=\linewidth]{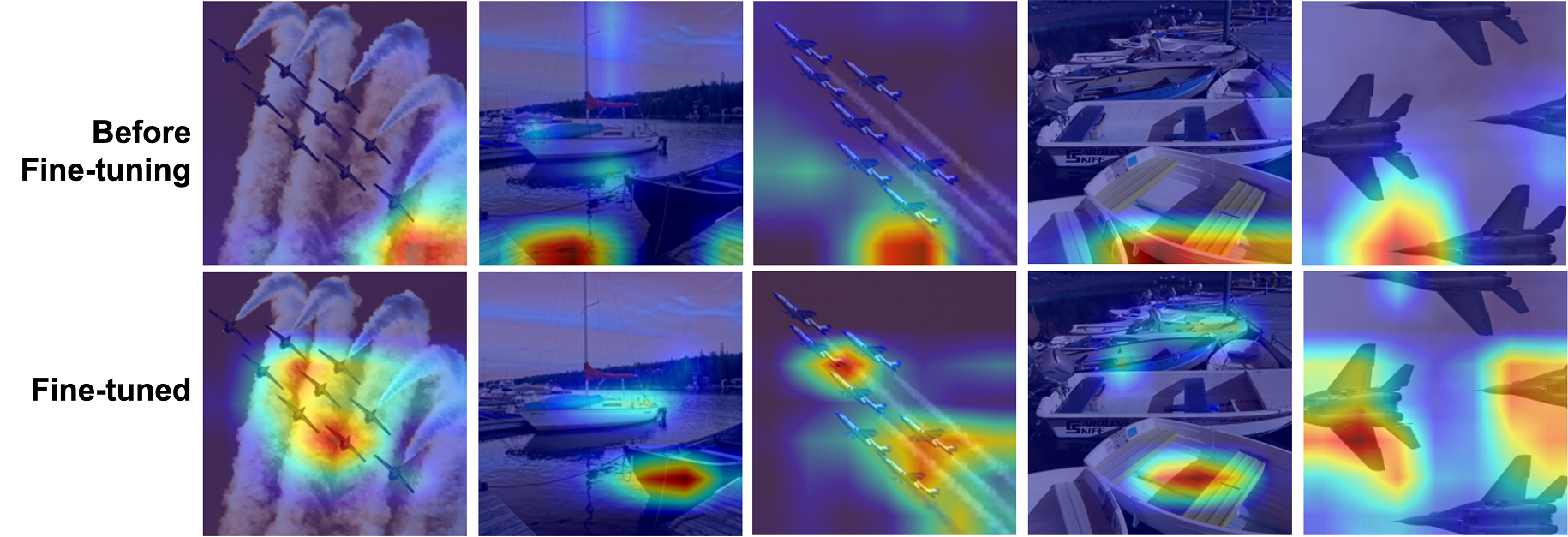}
    \caption{Some examples contain multiple features.}
    \label{fig:result_examples_multiple-features}
    \end{subfigure}
    \caption{Attention maps obtained by network before and after fine-tuning on different datasets. First and second rows show pretrained and fine-tuned results, respectively.}
    \label{fig:result_examples}
\end{figure}

In Figure~\ref{fig:result_examples}(a), we show attention maps visualized from images in COCO-2017 that were affected by dataset bias, as well as the attention map of the same images visualized using a fine-tuned network. Examples of AwA2 and CelebA are shown in Figures~\ref{fig:result_examples}(b) and \ref{fig:result_examples}(c), respectively. Although we only fine-tuned the pre-trained network for ten epochs, we discovered that the fine-tuned network can focus on the target object better than the pretrained network.

\paragraph{Compatibility for multiple features.}
In image classification problems, we encountered cases in which multiple target features are included in the same image. In this case, the DNN can respond to only one of the targets. We performed an experiment involving multiple target features appearing in the same image. We expect the fine-tuned network to respond to multiple target features in a picture simultaneously.
The results (Figure~\ref{fig:result_examples}(d)) show that this system can correct multiple feature targets in a single image.

\section{User Study}

We conducted a user study to evaluate the proposed system by comparing it with the baselines within the same duration. Our hypothesis is that the model fine-tuned using our proposed system can achieve higher accuracy and better feature representation and that the efficient framework yields higher-quality annotation.

\subsection{Participants}

To validate our hypothesis, we recruited laboratory students (from unrelated laboratories) with study or research experience in machine learning to perform the user study. Sixteen participants were invited to participate in the user evaluation process to evaluate click- and polygon-based user interfaces and active learning mechanisms, after which they were presented with a questionnaire. The age distribution of the participants was between 23 and 28 years, with a mean age of 24.5 years and an SD of 1.5 years. Among the participants, two participants stated that they were \emph{familiar with machine learning}, six participants \emph{using some machine learning algorithm in their research}, eight participants \emph{have experience in programming with some simple machine learning method}. Among the participants, 12 were male and 4 were female. The time spent was 3h, and each participant incurred \$30.

\subsection{Metrics}

We evaluated the system via a user study using both subjective and objective criteria. Two indicators were estimated for objective evaluations. The first was the classification accuracy of the network, which was used to demonstrate its performance after fine-tuning. We expect the accuracy of the pretrained network for the same classification task to be improved to a certain extent after sufficient fine-tuning. The second was the annotation time for each image, which significantly affects the crowdsourcing cost; furthermore, a longer operation time adversely affects the label quality.

The subjective evaluation was assessed using the user questionnaire detailed in Section~\ref{subsec:questionnaire}. We requested feedback from users based on the following criteria:

\begin{itemize}
    \item Effectiveness of the system (Q1, Q2, Q3 and Q4);
    \item Ability of the network to obtain a more reasonable attention map after fine-tuning (Q5 and Q6);
    \item Reliability of fine-tuned DNNs using the system (Q7).
\end{itemize}

\subsection{Baseline}

\begin{figure}[htp]
\centering
\begin{subfigure}{0.45\linewidth}
\centering
\includegraphics[width=\textwidth]{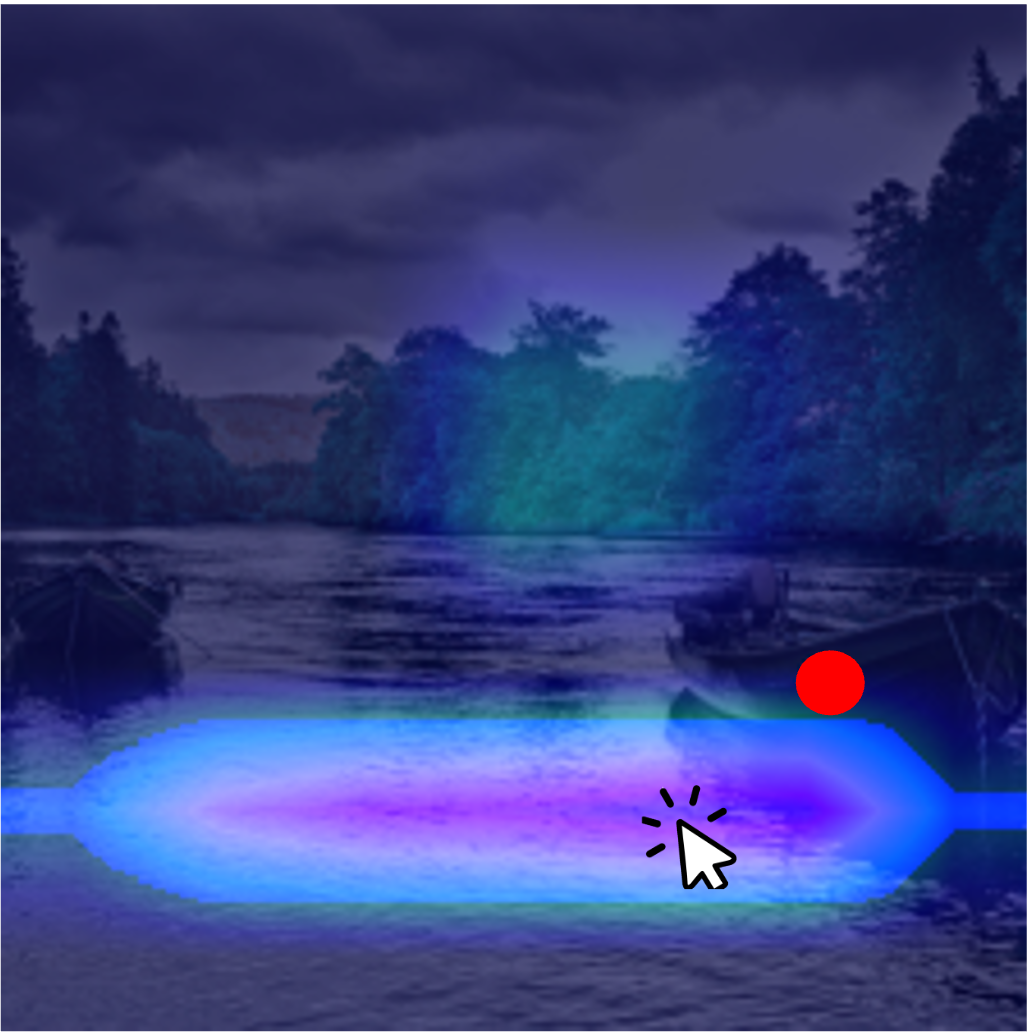}
\caption{Proposed click based approach.}
\label{fig:UI_click}
\end{subfigure}
\begin{subfigure}{0.45\linewidth}
\centering
\includegraphics[width=\textwidth]{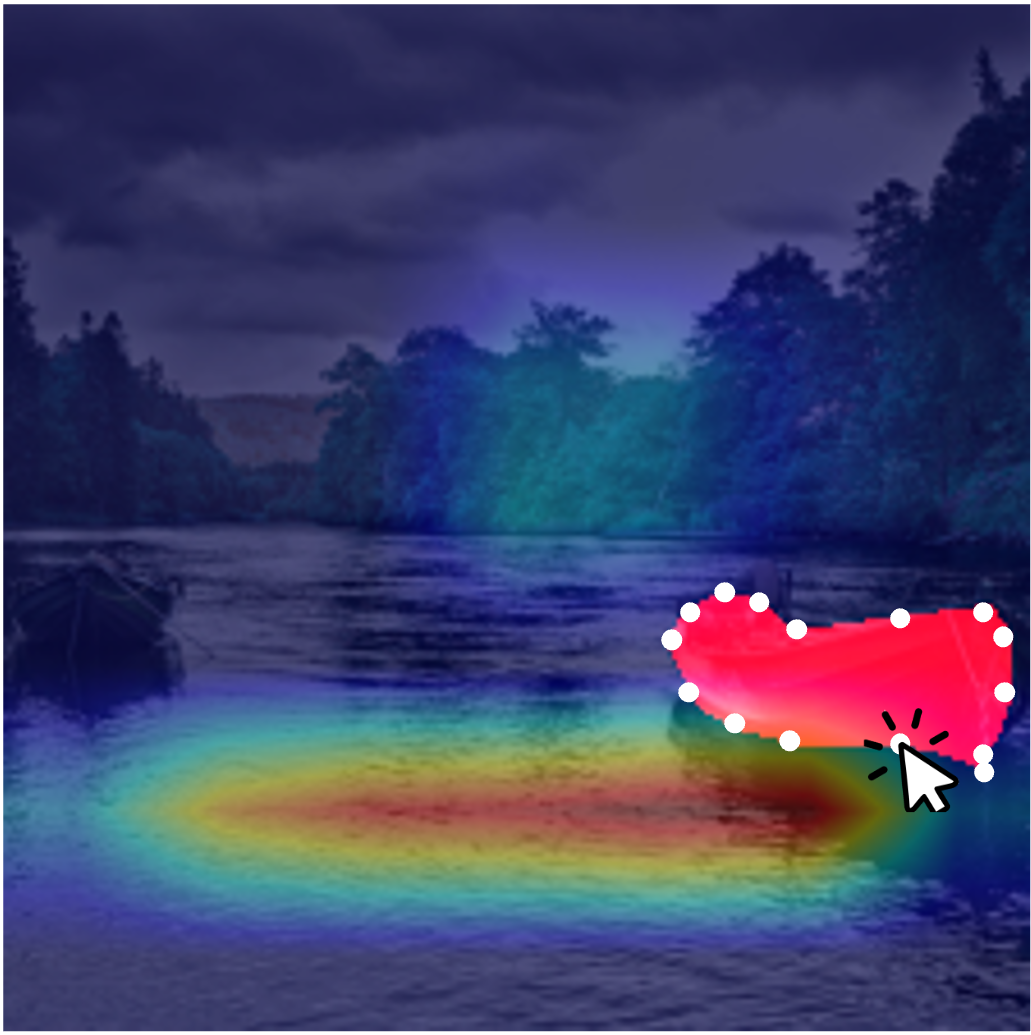}
\caption{Baseline polygon based approach.}
\label{fig:UI_segmentation}
\end{subfigure}
\caption{(a) Proposed click-based attention directing approach;(b) implemented polygon-based approach. Notably, the polygon-based approach does not share the function of the right and middle clicks with the click-based approach.}
\label{fig:ui_comparison}
\end{figure}

To evaluate our proposed system, we employed well-established methods, a polygon-based user interface for obtaining image annotations, and a random data selection strategy for image sampling as the baseline for comparison tests. Figure~\ref{fig:ui_comparison}(b) shows the traditional polygon-based annotation method used to select ROIs. In baseline situations, the user was only required to click on several vertices to draw the boundary of the target in the image. To investigate further, we conducted another comparison test with the polygon-based user interface and proposed attention-based data selection strategy.

\subsection{Procedure of User Study}

During the user study, the participants were instructed to sit in front of a desktop computer and complete four given tasks in a random order (within-subject design) \cite{field2002design}. Each task required 45 min to complete; hence, the total completion time was 180 min. The annotation systems were executed on a browser, and we explained the method to use the tools to the participants prior to the experiment. Although we expect the users to annotate all positive and negative regions using the left and right clicks, for example, left clicks for airplanes and right clicks for the sky or parking lots, in the actual user study, we allowed the participants to decide where the attention should or should not be.

We prepared two categories of image sets affected by dataset bias, i.e., ``airplane'' and ``boat'' as datasets. Airplanes typically appear in the sky or parking lots, whereas boats typically appear in water. During the user study, the number of candidates displayed on the UI at a time was set to 64, whereas the batch size, which refers to the number of images required for annotation during one cycle, was 256. Four test conditions were established: polygon-random, \textbf{polygon-active}, \textbf{click-random}, and \textbf{click-active}. \textbf{Polygon} implies polygon-based annotation interface (baseline), \textbf{click}, click-based annotation (proposed); \textbf{Random}, random-sampling (baseline); and \textbf{active}, attention-based data sampling (proposed). We randomized the order in which users performed the tasks and selected four orders; 16 participants were randomly assigned to the orders, with four participants in each order.

\subsection{Questionnaire}
\label{subsec:questionnaire}

After the users completed each task, we instructed them to complete a five-point Likert scale questionnaire to evaluate the usability of the user interface and system.
The contents of the questionnaire were as follows: 

\begin{enumerate}
    \item[Q1] Attention annotating with the click-based user interface was difficult.
    \item[Q2] Attention annotating with the polygon-based user interface was difficult.
    \item[Q3] It was difficult to concentrate on the tasks when using click-based user interface.
    \item[Q4] It was difficult to concentrate on the tasks when using polygon-based user interface.
    \item[Q5] The attention maps obtained by active learning was reasonable.
    \item[Q6] The attention maps obtained by random sampling was reasonable.
    \item[Q7] I am confident to use the fine-tuned DNNs in my further research or deployment.
    \item[Q8] Which fine-tune method do you prefer to use?
    \item[Q9] Why or some comments?
\end{enumerate}

\section{Results}

\begin{figure*}[h]
    \centering
    \begin{subfigure}{.49\linewidth}
      \centering
        \includegraphics[width=\linewidth]{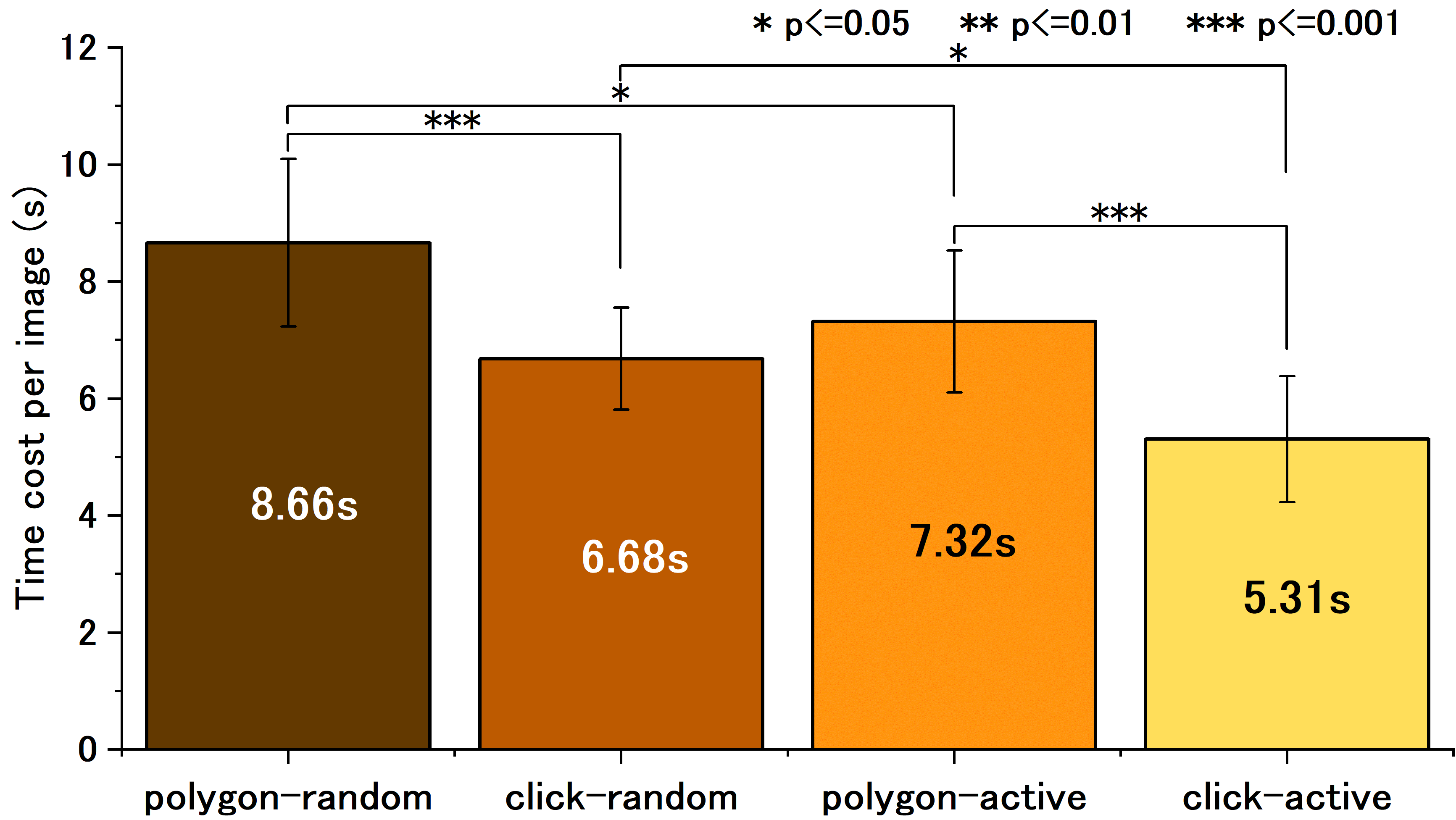}
        \caption{Average time cost on each image annotation.}
        \label{fig:user_study_time_cost}
    \end{subfigure}
    \begin{subfigure}{.49\linewidth}
      \centering
        \includegraphics[width=\linewidth]{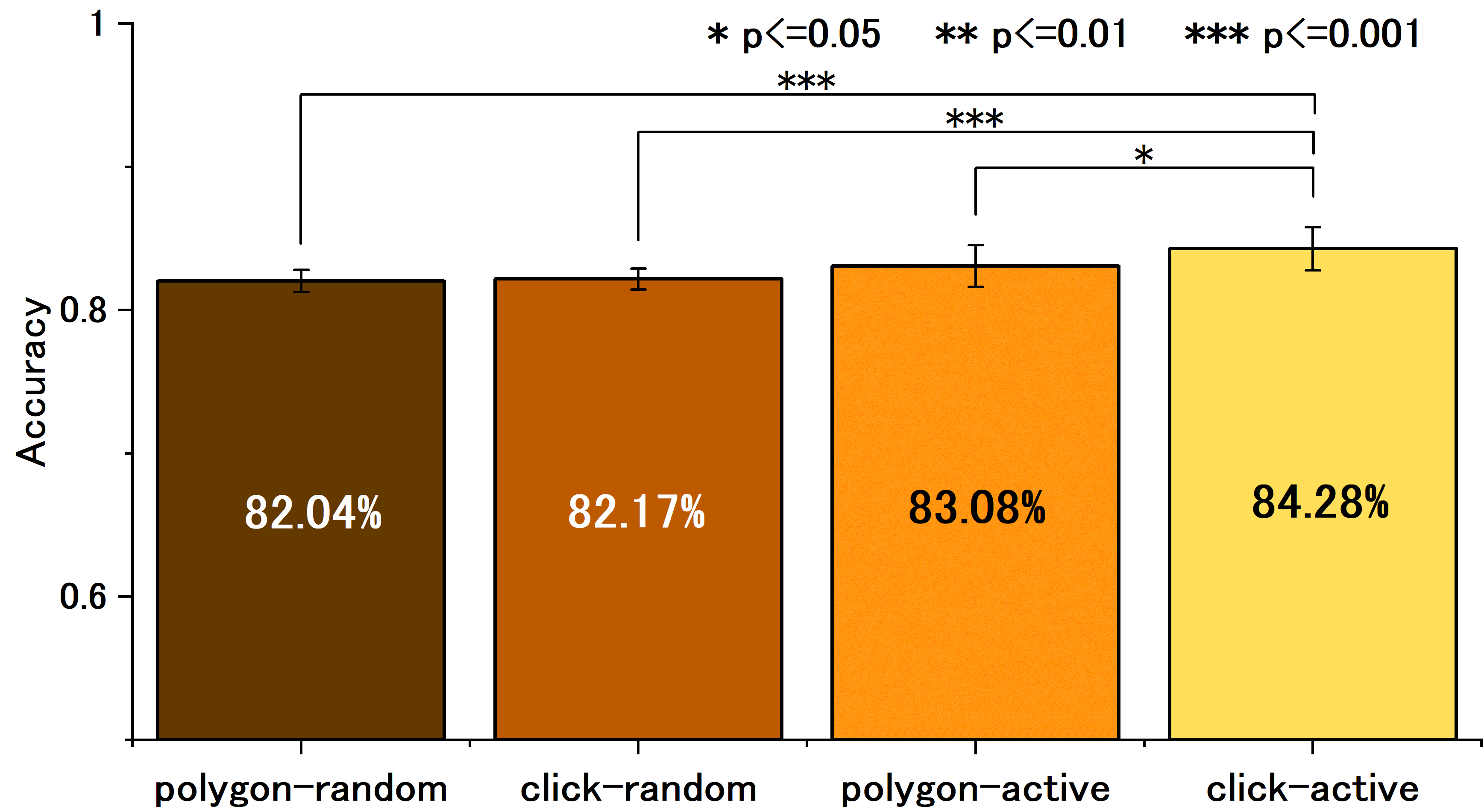}
        \caption{Average accuracy after fine-tuning.}
        \label{fig:user_study_accuracy}
    \end{subfigure}
    \caption{One-way analysis of variance and post-hoc tests were utilized for the data from the user study. In (a), we show the average time cost for each image annotation in the polygon-random task, with mean = 8.663 and SD = 1.431; click-random task, with mean = 6.680 and SD = 0.872; polygon-active task, with mean = 7.318 and SD = 1.211, and click-active task, with mean = 5.308 and SD = 1.076. In (b), we show the average accuracy before fine-tuning (baseline), i.e., 0.811. After fine-tuning, we obtained the average accuracy for the polygon-random task, with mean = 0.820 and SD = 0.008; click-random task, with mean = 0.822 and SD = 0.007; polygon-active task, with mean = 0.830 and SD = 0.014; and click-active task, with mean = 0.843 and SD = 0.015.}
    \label{fig:user_study}
\end{figure*}

\subsection{Time Cost}

Figure~\ref{fig:user_study}(a) shows that the participants spent 6.68 and 8.66s on average for the click-based and polygon-based approaches with a random strategy, respectively, whereas they spent 5.31 and 7.32s on average for those with an attention-based strategy, respectively, on each image requiring annotation. 
Switching from the polygon-based approach to the click-based approach with the random strategy afforded a 22.89\% decrease in time, whereas the time decrease was 27.46\% with the attention-based strategy. The results of this paired T-test show that utilizing two different methods resulted in a more significant difference (p<=0.001). This implies that the click-based approach significantly reduced the time required for attention annotation compared with the polygon-based approach. In addition, the results show that our proposed attention-based approach resulted in a more significant time cost decrease of 20.54\% (p<=0.05) when the same click-based annotation approach was used. This is because the approximate attention distribution of the images reduced the time required by the user in determining whether the network has extracted useful features from the target image.

\subsection{Accuracy after Fine-tuning}

Figure~\ref{fig:user_study}(b) shows the average accuracy after fine-tuning was performed. As the accuracy of our baseline model was 81.14\%, the participants obtained average accuracies of 82.04\% and 82.17\% after performing fine-tuning in the polygon-random and click-random conditions, respectively. The participants improved their accuracies by an average of 83.08\% and 84.29\% in the polygon-active and click-active conditions, respectively. We argue that our proposed active learning method significantly improves the efficiency of attention-directing tasks.

\subsection{Questionnaire}

\begin{figure*}[t]
    \centering
    \includegraphics[width=0.8\linewidth]{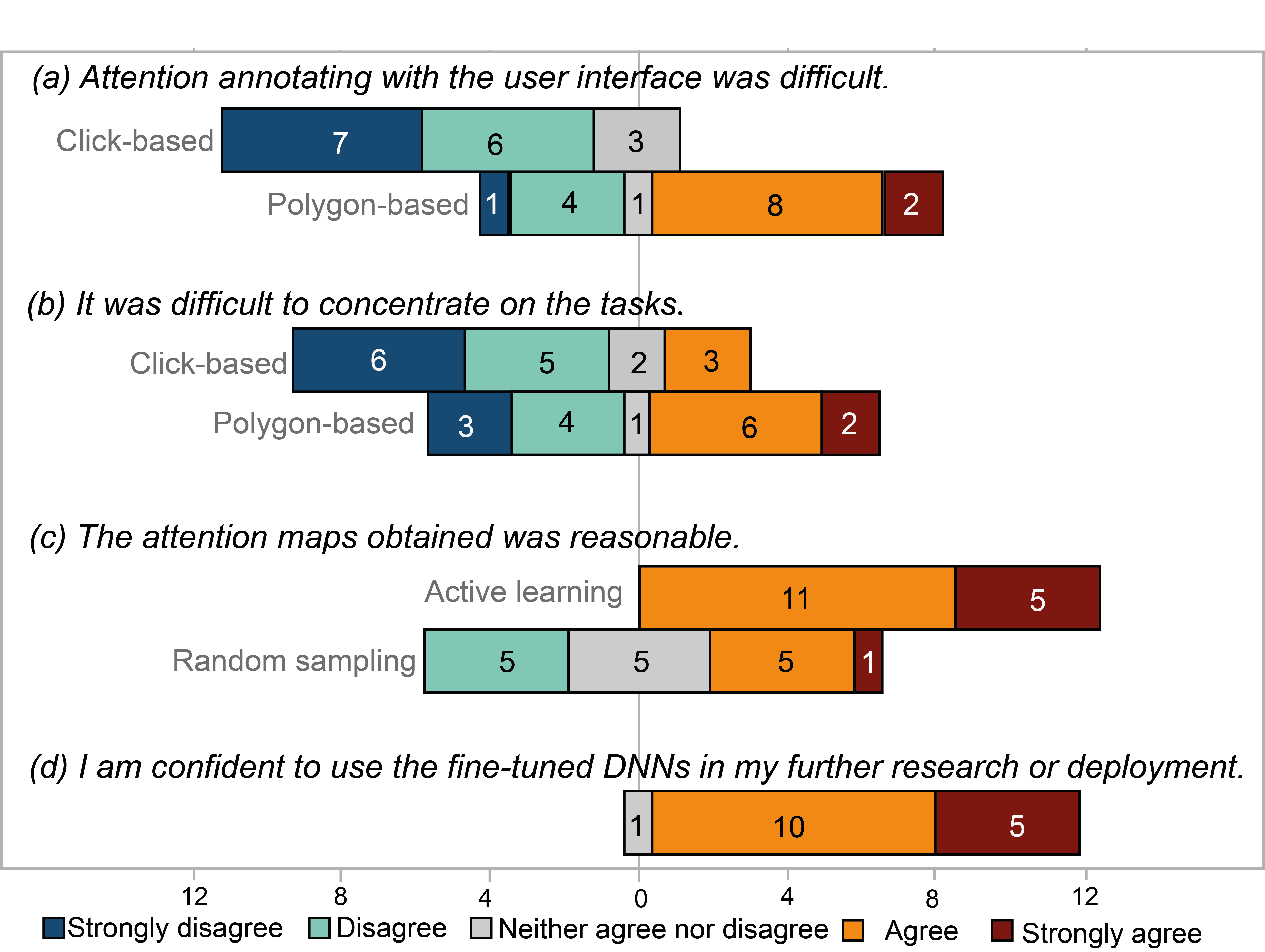}
    \caption{Results of user study questionnaire.}
    \label{fig:questionnaire}
\end{figure*}

Figure~\ref{fig:questionnaire} $(a)$ shows whether the tasks were difficult to accomplish using the click and polygon-based approaches. The results show that 62.5\% of the participants agreed (n = 8) or strongly agreed(n = 2) that using the polygon-based approach for the attention annotation task was difficult, whereas 81.25\% strongly disagreed (n = 7) or disagreed (n = 6) that using the click-based approach for the same task was difficult. This implies that using the click-based approach for the attention annotation task was burdenless to the participants. For example, one participant commented, \emph{"the click-based approach allowed me to focus on the target instead of the background or the boundary"}, and another participant commented, \emph{"the simple click approach rendered annotation easier and required less time."}.

We believe that the use of active learning for attention annotation is energy and concentration intensive. If a user's attention cannot be maintained for a long period, then the annotation quality may be deteriorated, and the attention optimization of the network will be less effective. Figure~\ref{fig:questionnaire} $(b)$ shows whether users can maintain a high level of concentration for a long period when using these two methods separately; 68.75\% of participants strongly disagreed (n = 6) or disagreed (n = 5) that they could maintain a high level of attention when using the click-based approach, whereas 43.75\% (n = 3 and n = 4, respectively) reported the same for with the polygon-based approach. One of the participants commented, \emph{"using the clicking method, I can focus on images that must be annotated, whereas the polygon approach only instructs me to select the boundaries of target objects in all the images, which is tiring."}

After the participants finished all the tasks, we showed them the results of the attention map visualization test for the fine-tuned network. Based on Figure~\ref{fig:questionnaire} $(c)$, for the same task time, all the participants strongly agreed (n = 5) or agreed (n = 11) that the results from the active learning were reasonable, whereas only 31.25\% of the participants agreed (n = 5) that the results of the random sampling were reasonable. Despite the short duration of the experiment, the active learning approach yielded relatively better results within a shorter duration because the random sampling yielded more valid data (which implies the necessity for user annotation) required for fine-tuning in the same amount of time.

As all the participants have experience in machine learning or are performing studies related to machine learning, we instructed them provide feedback regarding the reliability of the fine-tuned network. As shown in Figure~\ref{fig:questionnaire} $(d)$, most (n = 15) of the participants agreed that the results are reliable for deployment or further application.

The questionnaire results show that 75\% of the participants (n = 12) preferred the click-based approach, whereas only 25\% of the participants (n = 4) preferred the polygon-based approach. Among the participants who selected the click-based approach, most commented that the click-based approach rendered it easier to focus on selecting targets instead of the method to segment the target contours. Furthermore, they commented that such an approach was more efficient and allowed more useful data to be obtained within a shorter period. Most of the participants who preferred the click-based approach indicated that "It is easy
to conduct the annotation task with a simple operation". By contrast, those who preferred the polygon-based approach indicated that it selected the contour of the target object more accurately and was more conducive to fine-tuning. A participant who preferred the polygon-based approach commented, \emph{"I prefer the approach that yields a higher fine-tuning accuracy,"} whereas other participants who preferred the polygon-based approach argued, \emph{"I will not be able to maintain my concentration by performing boring, simple clicks for long durations."}
\section{Lessons Learned}

\subsection{Attention-based Data Sampling Strategy}

A data sampling strategy is critical for improving the effectiveness of active learning. Our attention-based strategy performs better than other typically used strategies owing to the following two reasons: First, to direct the attention of the DNNs, the attention map obviously provides more information than merely the possibility of classification. Second, unlike typical tasks, diverse images are more informative. Annotating images with typical attention maps is vital to our task.

Moreover, our attention-based strategy provides an additional benefit. The annotation time per image is significantly lower than that for the same user interface using the random strategy. By observing the participants and considering their feedback, we discovered that they can perform assessments and complete annotations faster when using our strategy. This is because the images selected by our strategy exhibit similar errors owing to their similar attention and texture features. By contrast, when using the random strategy, the users must consider various feature maps, which causes them to spend more time on decision-making.

\subsubsection{Efficiency and generalization}

One mutual comment from the participants who preferred the polygon-based approach was the following: \emph{"The polygon approach can frame objects with complex shapes, which allows a higher accuracy to be achieved under the same amount of annotations."}. Nonetheless, polygon-based annotations provide more exact shapes than click-based annotations. For the same number of annotated images, the network fine-tuned using exact attention shapes may yield better prediction results; however, the cost and time required for annotations are typically limited. Thus, owing to its efficiency, the click-based approach can be used to obtain more annotations and a better fine-tuned network. In addition, the complicated polygon-based approach is tedious to the users. As the operation time increased, the quality of the annotations became more difficult to maintain. Our click-based annotation can manage more tasks because drawing outlines with polygons is difficult if the target shape is ambiguous and unclear.

\subsubsection{Users are not oracles}

Some participants commented, \emph{"I was able to focus on images to be annotated and disregard unrelated items such as the background; however, I had trouble with the boundary of the target, which was frustrating"}, and \emph{"when I used the polygon-based annotation, I did not think that I was using my knowledge at all."} Users are people not oracles~\cite{amershi2014power}. The classical polygon-based approach does not require the user to focus on the output or errors of the network, nor does it require the user to consider the attention map. Compared with the classical polygon-based approach, our click-based approach is analogous to “teaching” the machine naturally. This allows users to participate more closely in the annotation process and provides more valuable experience and knowledge to the network.

\subsection{Further Functions}

Although the workload has been dramatically reduced, users tend to be inattentive and distracted when they cannot monitor the progress of their work. Thus, it is necessary to add functions to our user interface such that the users can be better informed of their annotation and fine-tuning progress. To maintain the annotations quality, more mechanisms can be added to our system to maintain the attention of users while they are labeling thousands of images, such as regular rest and attention tests. In addition, although we provided a feature that allows users to cancel all operations and re-annotate them, the users still commented that a redo feature that allows them to just undo the previous operation is necessary.

Our active learning and loss computation methods were designed under the hypothesis that the features learned by the network comprise two components (correct and co-occurrence features). Even if the expected results are achieved, they are based on empiricism and insight, not theory.

In the future, we would like to update the user interface and improve the system based on feedback from the user study. In particular, we intend to identify more natural interaction techniques that allow more human knowledge to be embedded into DNNs. We believe that the proposed system can be extended and applied to other multimodal data, in addition to images.

\section{Conclusion}

In this study, we proposed an efficient human-in-the-loop system to guide the attention of deep neural networks with a biased representation to focus on the region wherein classification decision evidence is located. We designed a user-friendly interface and a pointed active learning algorithm to significantly improve the performance and reduce the workload of the annotators. The results of our user study demonstrated the efficiency of our proposed system compared with the baseline method. We believe that our proposed system can promote the integration of interaction techniques and deep learning fields and render DNNs more convenient to apply for practitioners.

\begin{acks}
This work was supported by JST CREST Grant Number JPMJCR17A1, and JAIST Research Grant.
\end{acks}
\bibliographystyle{ACM-Reference-Format}
\bibliography{sample-base}


\begin{thebibliography}{69}


\ifx \showCODEN    \undefined \def \showCODEN     #1{\unskip}     \fi
\ifx \showDOI      \undefined \def \showDOI       #1{#1}\fi
\ifx \showISBNx    \undefined \def \showISBNx     #1{\unskip}     \fi
\ifx \showISBNxiii \undefined \def \showISBNxiii  #1{\unskip}     \fi
\ifx \showISSN     \undefined \def \showISSN      #1{\unskip}     \fi
\ifx \showLCCN     \undefined \def \showLCCN      #1{\unskip}     \fi
\ifx \shownote     \undefined \def \shownote      #1{#1}          \fi
\ifx \showarticletitle \undefined \def \showarticletitle #1{#1}   \fi
\ifx \showURL      \undefined \def \showURL       {\relax}        \fi
\providecommand\bibfield[2]{#2}
\providecommand\bibinfo[2]{#2}
\providecommand\natexlab[1]{#1}
\providecommand\showeprint[2][]{arXiv:#2}

\bibitem[fla(2022)]%
        {flask}
 \bibinfo{year}{2022}\natexlab{}.
\newblock \bibinfo{title}{The Python micro framework for building web
  applications.}
\newblock \bibinfo{howpublished}{\url{flask.palletsprojects.com}}.
\newblock


\bibitem[Adebayo et~al\mbox{.}(2018)]%
        {adebayo2018sanity}
\bibfield{author}{\bibinfo{person}{Julius Adebayo}, \bibinfo{person}{Justin
  Gilmer}, \bibinfo{person}{Michael Muelly}, \bibinfo{person}{Ian Goodfellow},
  \bibinfo{person}{Moritz Hardt}, {and} \bibinfo{person}{Been Kim}.}
  \bibinfo{year}{2018}\natexlab{}.
\newblock \showarticletitle{Sanity checks for saliency maps}.
\newblock \bibinfo{journal}{\emph{Advances in neural information processing
  systems}}  \bibinfo{volume}{31} (\bibinfo{year}{2018}).
\newblock


\bibitem[Agarwal et~al\mbox{.}(2020)]%
        {agarwal2020contextual}
\bibfield{author}{\bibinfo{person}{Sharat Agarwal}, \bibinfo{person}{Himanshu
  Arora}, \bibinfo{person}{Saket Anand}, {and} \bibinfo{person}{Chetan Arora}.}
  \bibinfo{year}{2020}\natexlab{}.
\newblock \showarticletitle{Contextual diversity for active learning}. In
  \bibinfo{booktitle}{\emph{European Conference on Computer Vision}}. Springer,
  \bibinfo{pages}{137--153}.
\newblock


\bibitem[Aggarwal et~al\mbox{.}(2020)]%
        {aggarwal2020active}
\bibfield{author}{\bibinfo{person}{Umang Aggarwal}, \bibinfo{person}{Adrian
  Popescu}, {and} \bibinfo{person}{C{\'e}line Hudelot}.}
  \bibinfo{year}{2020}\natexlab{}.
\newblock \showarticletitle{Active learning for imbalanced datasets}. In
  \bibinfo{booktitle}{\emph{Proceedings of the IEEE/CVF Winter Conference on
  Applications of Computer Vision}}. \bibinfo{pages}{1428--1437}.
\newblock


\bibitem[Agustsson et~al\mbox{.}(2019)]%
        {agustsson2019interactive}
\bibfield{author}{\bibinfo{person}{Eirikur Agustsson},
  \bibinfo{person}{Jasper~RR Uijlings}, {and} \bibinfo{person}{Vittorio
  Ferrari}.} \bibinfo{year}{2019}\natexlab{}.
\newblock \showarticletitle{Interactive full image segmentation by considering
  all regions jointly}. In \bibinfo{booktitle}{\emph{Proceedings of the
  IEEE/CVF Conference on Computer Vision and Pattern Recognition}}.
  \bibinfo{pages}{11622--11631}.
\newblock


\bibitem[Amershi et~al\mbox{.}(2014)]%
        {amershi2014power}
\bibfield{author}{\bibinfo{person}{Saleema Amershi}, \bibinfo{person}{Maya
  Cakmak}, \bibinfo{person}{William~Bradley Knox}, {and} \bibinfo{person}{Todd
  Kulesza}.} \bibinfo{year}{2014}\natexlab{}.
\newblock \showarticletitle{Power to the people: The role of humans in
  interactive machine learning}.
\newblock \bibinfo{journal}{\emph{Ai Magazine}} \bibinfo{volume}{35},
  \bibinfo{number}{4} (\bibinfo{year}{2014}), \bibinfo{pages}{105--120}.
\newblock


\bibitem[Arrieta et~al\mbox{.}(2020)]%
        {arrieta2020explainable}
\bibfield{author}{\bibinfo{person}{Alejandro~Barredo Arrieta},
  \bibinfo{person}{Natalia D{\'\i}az-Rodr{\'\i}guez}, \bibinfo{person}{Javier
  Del~Ser}, \bibinfo{person}{Adrien Bennetot}, \bibinfo{person}{Siham Tabik},
  \bibinfo{person}{Alberto Barbado}, \bibinfo{person}{Salvador Garc{\'\i}a},
  \bibinfo{person}{Sergio Gil-L{\'o}pez}, \bibinfo{person}{Daniel Molina},
  \bibinfo{person}{Richard Benjamins}, {et~al\mbox{.}}}
  \bibinfo{year}{2020}\natexlab{}.
\newblock \showarticletitle{Explainable Artificial Intelligence (XAI):
  Concepts, taxonomies, opportunities and challenges toward responsible AI}.
\newblock \bibinfo{journal}{\emph{Information fusion}}  \bibinfo{volume}{58}
  (\bibinfo{year}{2020}), \bibinfo{pages}{82--115}.
\newblock


\bibitem[Bahng et~al\mbox{.}(2020)]%
        {bahng2020learning}
\bibfield{author}{\bibinfo{person}{Hyojin Bahng}, \bibinfo{person}{Sanghyuk
  Chun}, \bibinfo{person}{Sangdoo Yun}, \bibinfo{person}{Jaegul Choo}, {and}
  \bibinfo{person}{Seong~Joon Oh}.} \bibinfo{year}{2020}\natexlab{}.
\newblock \showarticletitle{Learning de-biased representations with biased
  representations}. In \bibinfo{booktitle}{\emph{International Conference on
  Machine Learning}}. PMLR, \bibinfo{pages}{528--539}.
\newblock


\bibitem[Benenson et~al\mbox{.}(2019)]%
        {benenson2019large}
\bibfield{author}{\bibinfo{person}{Rodrigo Benenson}, \bibinfo{person}{Stefan
  Popov}, {and} \bibinfo{person}{Vittorio Ferrari}.}
  \bibinfo{year}{2019}\natexlab{}.
\newblock \showarticletitle{Large-scale interactive object segmentation with
  human annotators}. In \bibinfo{booktitle}{\emph{Proceedings of the IEEE/CVF
  Conference on Computer Vision and Pattern Recognition}}.
  \bibinfo{pages}{11700--11709}.
\newblock


\bibitem[Chiu et~al\mbox{.}(2020)]%
        {chiu2020human}
\bibfield{author}{\bibinfo{person}{Chia-Hsing Chiu}, \bibinfo{person}{Yuki
  Koyama}, \bibinfo{person}{Yu-Chi Lai}, \bibinfo{person}{Takeo Igarashi},
  {and} \bibinfo{person}{Yonghao Yue}.} \bibinfo{year}{2020}\natexlab{}.
\newblock \showarticletitle{Human-in-the-loop differential subspace search in
  high-dimensional latent space}.
\newblock \bibinfo{journal}{\emph{ACM Transactions on Graphics (TOG)}}
  \bibinfo{volume}{39}, \bibinfo{number}{4} (\bibinfo{year}{2020}),
  \bibinfo{pages}{85--1}.
\newblock


\bibitem[Dempster et~al\mbox{.}(1977)]%
        {dempster1977maximum}
\bibfield{author}{\bibinfo{person}{Arthur~P Dempster}, \bibinfo{person}{Nan~M
  Laird}, {and} \bibinfo{person}{Donald~B Rubin}.}
  \bibinfo{year}{1977}\natexlab{}.
\newblock \showarticletitle{Maximum likelihood from incomplete data via the EM
  algorithm}.
\newblock \bibinfo{journal}{\emph{Journal of the Royal Statistical Society:
  Series B (Methodological)}} \bibinfo{volume}{39}, \bibinfo{number}{1}
  (\bibinfo{year}{1977}), \bibinfo{pages}{1--22}.
\newblock


\bibitem[Felzenszwalb and Huttenlocher(2004)]%
        {felzenszwalb2004efficient}
\bibfield{author}{\bibinfo{person}{Pedro~F Felzenszwalb} {and}
  \bibinfo{person}{Daniel~P Huttenlocher}.} \bibinfo{year}{2004}\natexlab{}.
\newblock \showarticletitle{Efficient graph-based image segmentation}.
\newblock \bibinfo{journal}{\emph{International journal of computer vision}}
  \bibinfo{volume}{59}, \bibinfo{number}{2} (\bibinfo{year}{2004}),
  \bibinfo{pages}{167--181}.
\newblock


\bibitem[Field and Hole(2002)]%
        {field2002design}
\bibfield{author}{\bibinfo{person}{Andy Field} {and} \bibinfo{person}{Graham
  Hole}.} \bibinfo{year}{2002}\natexlab{}.
\newblock \bibinfo{booktitle}{\emph{How to design and report experiments}}.
\newblock \bibinfo{publisher}{Sage}.
\newblock


\bibitem[Fukui et~al\mbox{.}(2018)]%
        {fukui2018attention}
\bibfield{author}{\bibinfo{person}{H. Fukui}, \bibinfo{person}{T. Hirakawa},
  \bibinfo{person}{T. Yamashita}, {and} \bibinfo{person}{H. Fujiyoshi}.}
  \bibinfo{year}{2018}\natexlab{}.
\newblock \showarticletitle{Attention Branch Network: Learning of Attention
  Mechanism for Visual Explanation}.
\newblock \bibinfo{journal}{\emph{arXiv}} (\bibinfo{year}{2018}).
\newblock


\bibitem[Gudovskiy et~al\mbox{.}(2020)]%
        {gudovskiy2020deep}
\bibfield{author}{\bibinfo{person}{Denis Gudovskiy}, \bibinfo{person}{Alec
  Hodgkinson}, \bibinfo{person}{Takuya Yamaguchi}, {and}
  \bibinfo{person}{Sotaro Tsukizawa}.} \bibinfo{year}{2020}\natexlab{}.
\newblock \showarticletitle{Deep active learning for biased datasets via fisher
  kernel self-supervision}. In \bibinfo{booktitle}{\emph{Proceedings of the
  IEEE/CVF Conference on Computer Vision and Pattern Recognition}}.
  \bibinfo{pages}{9041--9049}.
\newblock


\bibitem[Guidotti et~al\mbox{.}(2018)]%
        {guidotti2018survey}
\bibfield{author}{\bibinfo{person}{Riccardo Guidotti}, \bibinfo{person}{Anna
  Monreale}, \bibinfo{person}{Salvatore Ruggieri}, \bibinfo{person}{Franco
  Turini}, \bibinfo{person}{Fosca Giannotti}, {and} \bibinfo{person}{Dino
  Pedreschi}.} \bibinfo{year}{2018}\natexlab{}.
\newblock \showarticletitle{A survey of methods for explaining black box
  models}.
\newblock \bibinfo{journal}{\emph{ACM computing surveys (CSUR)}}
  \bibinfo{volume}{51}, \bibinfo{number}{5} (\bibinfo{year}{2018}),
  \bibinfo{pages}{1--42}.
\newblock


\bibitem[He et~al\mbox{.}(2016)]%
        {he2016deep}
\bibfield{author}{\bibinfo{person}{Kaiming He}, \bibinfo{person}{Xiangyu
  Zhang}, \bibinfo{person}{Shaoqing Ren}, {and} \bibinfo{person}{Jian Sun}.}
  \bibinfo{year}{2016}\natexlab{}.
\newblock \showarticletitle{Deep residual learning for image recognition}. In
  \bibinfo{booktitle}{\emph{Proceedings of the IEEE conference on computer
  vision and pattern recognition}}. \bibinfo{pages}{770--778}.
\newblock


\bibitem[Hendricks et~al\mbox{.}(2018)]%
        {hendricks2018women}
\bibfield{author}{\bibinfo{person}{Lisa~Anne Hendricks},
  \bibinfo{person}{Kaylee Burns}, \bibinfo{person}{Kate Saenko},
  \bibinfo{person}{Trevor Darrell}, {and} \bibinfo{person}{Anna Rohrbach}.}
  \bibinfo{year}{2018}\natexlab{}.
\newblock \showarticletitle{Women Also Snowboard: Overcoming Bias in Captioning
  Models}. In \bibinfo{booktitle}{\emph{European Conference on Computer
  Vision}}. Springer, \bibinfo{pages}{793--811}.
\newblock


\bibitem[Hilgard et~al\mbox{.}(2021)]%
        {hilgard2021learning}
\bibfield{author}{\bibinfo{person}{Sophie Hilgard}, \bibinfo{person}{Nir
  Rosenfeld}, \bibinfo{person}{Mahzarin~R Banaji}, \bibinfo{person}{Jack Cao},
  {and} \bibinfo{person}{David Parkes}.} \bibinfo{year}{2021}\natexlab{}.
\newblock \showarticletitle{Learning representations by humans, for humans}. In
  \bibinfo{booktitle}{\emph{International Conference on Machine Learning}}.
  PMLR, \bibinfo{pages}{4227--4238}.
\newblock


\bibitem[Khosla et~al\mbox{.}(2012)]%
        {khosla2012undoing}
\bibfield{author}{\bibinfo{person}{Aditya Khosla}, \bibinfo{person}{Tinghui
  Zhou}, \bibinfo{person}{Tomasz Malisiewicz}, \bibinfo{person}{Alexei~A
  Efros}, {and} \bibinfo{person}{Antonio Torralba}.}
  \bibinfo{year}{2012}\natexlab{}.
\newblock \showarticletitle{Undoing the damage of dataset bias}. In
  \bibinfo{booktitle}{\emph{European Conference on Computer Vision}}. Springer,
  \bibinfo{pages}{158--171}.
\newblock


\bibitem[Kim et~al\mbox{.}(2019)]%
        {kim2019learning}
\bibfield{author}{\bibinfo{person}{Byungju Kim}, \bibinfo{person}{Hyunwoo Kim},
  \bibinfo{person}{Kyungsu Kim}, \bibinfo{person}{Sungjin Kim}, {and}
  \bibinfo{person}{Junmo Kim}.} \bibinfo{year}{2019}\natexlab{}.
\newblock \showarticletitle{Learning not to learn: Training deep neural
  networks with biased data}. In \bibinfo{booktitle}{\emph{Proceedings of the
  IEEE/CVF Conference on Computer Vision and Pattern Recognition}}.
  \bibinfo{pages}{9012--9020}.
\newblock


\bibitem[Koh and Liang(2017)]%
        {koh2017understanding}
\bibfield{author}{\bibinfo{person}{Pang~Wei Koh} {and} \bibinfo{person}{Percy
  Liang}.} \bibinfo{year}{2017}\natexlab{}.
\newblock \showarticletitle{Understanding Black-box Predictions via Influence
  Functions}. In \bibinfo{booktitle}{\emph{International Conference on Machine
  Learning}}. \bibinfo{pages}{1885--1894}.
\newblock


\bibitem[Koyama et~al\mbox{.}(2020)]%
        {koyama2020sequential}
\bibfield{author}{\bibinfo{person}{Yuki Koyama}, \bibinfo{person}{Issei Sato},
  {and} \bibinfo{person}{Masataka Goto}.} \bibinfo{year}{2020}\natexlab{}.
\newblock \showarticletitle{Sequential gallery for interactive visual design
  optimization}.
\newblock \bibinfo{journal}{\emph{ACM Transactions on Graphics (TOG)}}
  \bibinfo{volume}{39}, \bibinfo{number}{4} (\bibinfo{year}{2020}),
  \bibinfo{pages}{88--1}.
\newblock


\bibitem[Koyama et~al\mbox{.}(2017)]%
        {koyama2017sequential}
\bibfield{author}{\bibinfo{person}{Yuki Koyama}, \bibinfo{person}{Issei Sato},
  \bibinfo{person}{Daisuke Sakamoto}, {and} \bibinfo{person}{Takeo Igarashi}.}
  \bibinfo{year}{2017}\natexlab{}.
\newblock \showarticletitle{Sequential line search for efficient visual design
  optimization by crowds}.
\newblock \bibinfo{journal}{\emph{ACM Transactions on Graphics (TOG)}}
  \bibinfo{volume}{36}, \bibinfo{number}{4} (\bibinfo{year}{2017}),
  \bibinfo{pages}{1--11}.
\newblock


\bibitem[Krizhevsky et~al\mbox{.}(2012)]%
        {krizhevsky2012imagenet}
\bibfield{author}{\bibinfo{person}{Alex Krizhevsky}, \bibinfo{person}{Ilya
  Sutskever}, {and} \bibinfo{person}{Geoffrey~E Hinton}.}
  \bibinfo{year}{2012}\natexlab{}.
\newblock \showarticletitle{Imagenet classification with deep convolutional
  neural networks}. In \bibinfo{booktitle}{\emph{Advances in neural information
  processing systems}}. \bibinfo{pages}{1097--1105}.
\newblock


\bibitem[Li et~al\mbox{.}(2018)]%
        {li2018tell}
\bibfield{author}{\bibinfo{person}{K. Li}, \bibinfo{person}{Z. Wu},
  \bibinfo{person}{K. Peng}, \bibinfo{person}{J. Ernst}, {and}
  \bibinfo{person}{Y. Fu}.} \bibinfo{year}{2018}\natexlab{}.
\newblock \showarticletitle{Tell me where to look: Guided attention inference
  network}. In \bibinfo{booktitle}{\emph{ICCV}}.
\newblock


\bibitem[Li and Vasconcelos(2019)]%
        {li2019repair}
\bibfield{author}{\bibinfo{person}{Yi Li} {and} \bibinfo{person}{Nuno
  Vasconcelos}.} \bibinfo{year}{2019}\natexlab{}.
\newblock \showarticletitle{Repair: Removing representation bias by dataset
  resampling}. In \bibinfo{booktitle}{\emph{Proceedings of the IEEE/CVF
  Conference on Computer Vision and Pattern Recognition}}.
  \bibinfo{pages}{9572--9581}.
\newblock


\bibitem[Lin et~al\mbox{.}(2015)]%
        {lin2015microsoft}
\bibfield{author}{\bibinfo{person}{Tsung-Yi Lin}, \bibinfo{person}{Michael
  Maire}, \bibinfo{person}{Serge Belongie}, \bibinfo{person}{Lubomir Bourdev},
  \bibinfo{person}{Ross Girshick}, \bibinfo{person}{James Hays},
  \bibinfo{person}{Pietro Perona}, \bibinfo{person}{Deva Ramanan},
  \bibinfo{person}{C.~Lawrence Zitnick}, {and} \bibinfo{person}{Piotr
  Dollár}.} \bibinfo{year}{2015}\natexlab{}.
\newblock \bibinfo{title}{Microsoft COCO: Common Objects in Context}.
\newblock
\newblock
\showeprint[arxiv]{1405.0312}~[cs.CV]


\bibitem[Liu et~al\mbox{.}(2017)]%
        {LIU201748}
\bibfield{author}{\bibinfo{person}{Shixia Liu}, \bibinfo{person}{Xiting Wang},
  \bibinfo{person}{Mengchen Liu}, {and} \bibinfo{person}{Jun Zhu}.}
  \bibinfo{year}{2017}\natexlab{}.
\newblock \showarticletitle{Towards better analysis of machine learning models:
  A visual analytics perspective}.
\newblock \bibinfo{journal}{\emph{Visual Informatics}} \bibinfo{volume}{1},
  \bibinfo{number}{1} (\bibinfo{year}{2017}), \bibinfo{pages}{48 -- 56}.
\newblock
\showISSN{2468-502X}
\urldef\tempurl%
\url{https://doi.org/10.1016/j.visinf.2017.01.006}
\showDOI{\tempurl}


\bibitem[Liu et~al\mbox{.}(2015)]%
        {liu2015faceattributes}
\bibfield{author}{\bibinfo{person}{Ziwei Liu}, \bibinfo{person}{Ping Luo},
  \bibinfo{person}{Xiaogang Wang}, {and} \bibinfo{person}{Xiaoou Tang}.}
  \bibinfo{year}{2015}\natexlab{}.
\newblock \showarticletitle{Deep Learning Face Attributes in the Wild}. In
  \bibinfo{booktitle}{\emph{Proceedings of International Conference on Computer
  Vision (ICCV)}}.
\newblock


\bibitem[Liu et~al\mbox{.}(2019)]%
        {liu2019deep}
\bibfield{author}{\bibinfo{person}{Zimo Liu}, \bibinfo{person}{Jingya Wang},
  \bibinfo{person}{Shaogang Gong}, \bibinfo{person}{Huchuan Lu}, {and}
  \bibinfo{person}{Dacheng Tao}.} \bibinfo{year}{2019}\natexlab{}.
\newblock \showarticletitle{Deep reinforcement active learning for
  human-in-the-loop person re-identification}. In
  \bibinfo{booktitle}{\emph{Proceedings of the IEEE/CVF International
  Conference on Computer Vision}}. \bibinfo{pages}{6122--6131}.
\newblock


\bibitem[Luo et~al\mbox{.}(2013)]%
        {NIPS2013_b6f0479a}
\bibfield{author}{\bibinfo{person}{Wenjie Luo}, \bibinfo{person}{Alex Schwing},
  {and} \bibinfo{person}{Raquel Urtasun}.} \bibinfo{year}{2013}\natexlab{}.
\newblock \showarticletitle{Latent Structured Active Learning}. In
  \bibinfo{booktitle}{\emph{Advances in Neural Information Processing
  Systems}}, \bibfield{editor}{\bibinfo{person}{C.~J.~C. Burges},
  \bibinfo{person}{L.~Bottou}, \bibinfo{person}{M.~Welling},
  \bibinfo{person}{Z.~Ghahramani}, {and} \bibinfo{person}{K.~Q. Weinberger}}
  (Eds.), Vol.~\bibinfo{volume}{26}. \bibinfo{publisher}{Curran Associates,
  Inc.}
\newblock
\urldef\tempurl%
\url{https://proceedings.neurips.cc/paper/2013/file/b6f0479ae87d244975439c6124592772-Paper.pdf}
\showURL{%
\tempurl}


\bibitem[Majumder and Yao(2019)]%
        {majumder2019content}
\bibfield{author}{\bibinfo{person}{Soumajit Majumder} {and}
  \bibinfo{person}{Angela Yao}.} \bibinfo{year}{2019}\natexlab{}.
\newblock \showarticletitle{Content-aware multi-level guidance for interactive
  instance segmentation}. In \bibinfo{booktitle}{\emph{Proceedings of the
  IEEE/CVF Conference on Computer Vision and Pattern Recognition}}.
  \bibinfo{pages}{11602--11611}.
\newblock


\bibitem[Maksai and Fua(2019)]%
        {maksai2019eliminating}
\bibfield{author}{\bibinfo{person}{Andrii Maksai} {and} \bibinfo{person}{Pascal
  Fua}.} \bibinfo{year}{2019}\natexlab{}.
\newblock \showarticletitle{Eliminating exposure bias and metric mismatch in
  multiple object tracking}. In \bibinfo{booktitle}{\emph{Proceedings of the
  IEEE/CVF Conference on Computer Vision and Pattern Recognition}}.
  \bibinfo{pages}{4639--4648}.
\newblock


\bibitem[Mishra and Rzeszotarski(2021)]%
        {mishra2021designing}
\bibfield{author}{\bibinfo{person}{Swati Mishra} {and}
  \bibinfo{person}{Jeffrey~M Rzeszotarski}.} \bibinfo{year}{2021}\natexlab{}.
\newblock \showarticletitle{Designing Interactive Transfer Learning Tools for
  ML Non-Experts}. In \bibinfo{booktitle}{\emph{Proceedings of the 2021 CHI
  Conference on Human Factors in Computing Systems}}. \bibinfo{pages}{1--15}.
\newblock


\bibitem[Murphy(2012)]%
        {murphy2012machine}
\bibfield{author}{\bibinfo{person}{Kevin~P Murphy}.}
  \bibinfo{year}{2012}\natexlab{}.
\newblock \bibinfo{booktitle}{\emph{Machine learning: a probabilistic
  perspective}}.
\newblock \bibinfo{publisher}{MIT press}.
\newblock


\bibitem[Nakano et~al\mbox{.}(2020)]%
        {nakano2020interactive}
\bibfield{author}{\bibinfo{person}{Tomoyasu Nakano}, \bibinfo{person}{Yuki
  Koyama}, \bibinfo{person}{Masahiro Hamasaki}, {and} \bibinfo{person}{Masataka
  Goto}.} \bibinfo{year}{2020}\natexlab{}.
\newblock \showarticletitle{Interactive deep singing-voice separation based on
  human-in-the-loop adaptation}. In \bibinfo{booktitle}{\emph{Proceedings of
  the 25th International Conference on Intelligent User Interfaces}}.
  \bibinfo{pages}{78--82}.
\newblock


\bibitem[Paszke et~al\mbox{.}(2019)]%
        {NEURIPS2019_9015}
\bibfield{author}{\bibinfo{person}{Adam Paszke}, \bibinfo{person}{Sam Gross},
  \bibinfo{person}{Francisco Massa}, \bibinfo{person}{Adam Lerer},
  \bibinfo{person}{James Bradbury}, \bibinfo{person}{Gregory Chanan},
  \bibinfo{person}{Trevor Killeen}, \bibinfo{person}{Zeming Lin},
  \bibinfo{person}{Natalia Gimelshein}, \bibinfo{person}{Luca Antiga},
  \bibinfo{person}{Alban Desmaison}, \bibinfo{person}{Andreas Kopf},
  \bibinfo{person}{Edward Yang}, \bibinfo{person}{Zachary DeVito},
  \bibinfo{person}{Martin Raison}, \bibinfo{person}{Alykhan Tejani},
  \bibinfo{person}{Sasank Chilamkurthy}, \bibinfo{person}{Benoit Steiner},
  \bibinfo{person}{Lu Fang}, \bibinfo{person}{Junjie Bai}, {and}
  \bibinfo{person}{Soumith Chintala}.} \bibinfo{year}{2019}\natexlab{}.
\newblock \showarticletitle{PyTorch: An Imperative Style, High-Performance Deep
  Learning Library}.
\newblock In \bibinfo{booktitle}{\emph{Advances in Neural Information
  Processing Systems 32}}, \bibfield{editor}{\bibinfo{person}{H.~Wallach},
  \bibinfo{person}{H.~Larochelle}, \bibinfo{person}{A.~Beygelzimer},
  \bibinfo{person}{F.~d\textquotesingle Alch\'{e}-Buc},
  \bibinfo{person}{E.~Fox}, {and} \bibinfo{person}{R.~Garnett}} (Eds.).
  \bibinfo{publisher}{Curran Associates, Inc.}, \bibinfo{pages}{8024--8035}.
\newblock
\urldef\tempurl%
\url{http://papers.neurips.cc/paper/9015-pytorch-an-imperative-style-high-performance-deep-learning-library.pdf}
\showURL{%
\tempurl}


\bibitem[Ribeiro et~al\mbox{.}(2016)]%
        {ribeiro2016should}
\bibfield{author}{\bibinfo{person}{Marco~Tulio Ribeiro},
  \bibinfo{person}{Sameer Singh}, {and} \bibinfo{person}{Carlos Guestrin}.}
  \bibinfo{year}{2016}\natexlab{}.
\newblock \showarticletitle{" Why should i trust you?" Explaining the
  predictions of any classifier}. In \bibinfo{booktitle}{\emph{Proceedings of
  the 22nd ACM SIGKDD international conference on knowledge discovery and data
  mining}}. \bibinfo{pages}{1135--1144}.
\newblock


\bibitem[Selvaraju et~al\mbox{.}(2017)]%
        {selvaraju2017grad}
\bibfield{author}{\bibinfo{person}{Ramprasaath~R Selvaraju},
  \bibinfo{person}{Michael Cogswell}, \bibinfo{person}{Abhishek Das},
  \bibinfo{person}{Ramakrishna Vedantam}, \bibinfo{person}{Devi Parikh},
  \bibinfo{person}{Dhruv Batra}, {et~al\mbox{.}}}
  \bibinfo{year}{2017}\natexlab{}.
\newblock \showarticletitle{Grad-CAM: Visual Explanations from Deep Networks
  via Gradient-Based Localization.}. In \bibinfo{booktitle}{\emph{ICCV}}.
  \bibinfo{pages}{618--626}.
\newblock


\bibitem[Sener and Savarese(2017)]%
        {sener2017active}
\bibfield{author}{\bibinfo{person}{Ozan Sener} {and} \bibinfo{person}{Silvio
  Savarese}.} \bibinfo{year}{2017}\natexlab{}.
\newblock \showarticletitle{Active learning for convolutional neural networks:
  A core-set approach}.
\newblock \bibinfo{journal}{\emph{arXiv preprint arXiv:1708.00489}}
  (\bibinfo{year}{2017}).
\newblock


\bibitem[Sener and Savarese(2018)]%
        {sener2018active}
\bibfield{author}{\bibinfo{person}{Ozan Sener} {and} \bibinfo{person}{Silvio
  Savarese}.} \bibinfo{year}{2018}\natexlab{}.
\newblock \showarticletitle{Active Learning for Convolutional Neural Networks:
  A Core-Set Approach}. In \bibinfo{booktitle}{\emph{International Conference
  on Learning Representations}}.
\newblock
\urldef\tempurl%
\url{https://openreview.net/forum?id=H1aIuk-RW}
\showURL{%
\tempurl}


\bibitem[Settles(2009)]%
        {settles2009active}
\bibfield{author}{\bibinfo{person}{Burr Settles}.}
  \bibinfo{year}{2009}\natexlab{}.
\newblock \showarticletitle{Active learning literature survey}.
\newblock  (\bibinfo{year}{2009}).
\newblock


\bibitem[Settles and Craven(2008)]%
        {10.5555/1613715.1613855}
\bibfield{author}{\bibinfo{person}{Burr Settles} {and} \bibinfo{person}{Mark
  Craven}.} \bibinfo{year}{2008}\natexlab{}.
\newblock \showarticletitle{An Analysis of Active Learning Strategies for
  Sequence Labeling Tasks}. In \bibinfo{booktitle}{\emph{Proceedings of the
  Conference on Empirical Methods in Natural Language Processing}} (Honolulu,
  Hawaii) \emph{(\bibinfo{series}{EMNLP '08})}. \bibinfo{publisher}{Association
  for Computational Linguistics}, \bibinfo{address}{USA},
  \bibinfo{pages}{1070–1079}.
\newblock


\bibitem[Shen et~al\mbox{.}(2021)]%
        {shen2021human}
\bibfield{author}{\bibinfo{person}{Haifeng Shen}, \bibinfo{person}{Kewen Liao},
  \bibinfo{person}{Zhibin Liao}, \bibinfo{person}{Job Doornberg},
  \bibinfo{person}{Maoying Qiao}, \bibinfo{person}{Anton Van Den~Hengel}, {and}
  \bibinfo{person}{Johan~W Verjans}.} \bibinfo{year}{2021}\natexlab{}.
\newblock \showarticletitle{Human-AI Interactive and Continuous Sensemaking: A
  Case Study of Image Classification using Scribble Attention Maps}. In
  \bibinfo{booktitle}{\emph{Extended Abstracts of the 2021 CHI Conference on
  Human Factors in Computing Systems}}. \bibinfo{pages}{1--8}.
\newblock


\bibitem[Shimizu et~al\mbox{.}(2020)]%
        {shimizu2020design}
\bibfield{author}{\bibinfo{person}{Evan Shimizu}, \bibinfo{person}{Matthew
  Fisher}, \bibinfo{person}{Sylvain Paris}, \bibinfo{person}{James McCann},
  {and} \bibinfo{person}{Kayvon Fatahalian}.} \bibinfo{year}{2020}\natexlab{}.
\newblock \showarticletitle{Design Adjectives: A Framework for Interactive
  Model-Guided Exploration of Parameterized Design Spaces}. In
  \bibinfo{booktitle}{\emph{Proceedings of the 33rd Annual ACM Symposium on
  User Interface Software and Technology}}. \bibinfo{pages}{261--278}.
\newblock


\bibitem[Siddiqui et~al\mbox{.}(2020)]%
        {siddiqui2020viewal}
\bibfield{author}{\bibinfo{person}{Yawar Siddiqui}, \bibinfo{person}{Julien
  Valentin}, {and} \bibinfo{person}{Matthias Nie{\ss}ner}.}
  \bibinfo{year}{2020}\natexlab{}.
\newblock \showarticletitle{Viewal: Active learning with viewpoint entropy for
  semantic segmentation}. In \bibinfo{booktitle}{\emph{Proceedings of the
  IEEE/CVF Conference on Computer Vision and Pattern Recognition}}.
  \bibinfo{pages}{9433--9443}.
\newblock


\bibitem[Silva and Gombolay(2021)]%
        {silva2021encoding}
\bibfield{author}{\bibinfo{person}{Andrew Silva} {and} \bibinfo{person}{Matthew
  Gombolay}.} \bibinfo{year}{2021}\natexlab{}.
\newblock \showarticletitle{Encoding Human Domain Knowledge to Warm Start
  Reinforcement Learning}. In \bibinfo{booktitle}{\emph{Proceedings of the AAAI
  Conference on Artificial Intelligence}}, Vol.~\bibinfo{volume}{35}.
  \bibinfo{pages}{5042--5050}.
\newblock


\bibitem[Simonyan et~al\mbox{.}(2013)]%
        {Simonyan2013DeepIC}
\bibfield{author}{\bibinfo{person}{Karen Simonyan}, \bibinfo{person}{Andrea
  Vedaldi}, {and} \bibinfo{person}{Andrew Zisserman}.}
  \bibinfo{year}{2013}\natexlab{}.
\newblock \showarticletitle{Deep Inside Convolutional Networks: Visualising
  Image Classification Models and Saliency Maps}.
\newblock \bibinfo{journal}{\emph{CoRR}}  \bibinfo{volume}{abs/1312.6034}
  (\bibinfo{year}{2013}).
\newblock


\bibitem[Sinha et~al\mbox{.}(2019)]%
        {sinha2019variational}
\bibfield{author}{\bibinfo{person}{Samarth Sinha}, \bibinfo{person}{Sayna
  Ebrahimi}, {and} \bibinfo{person}{Trevor Darrell}.}
  \bibinfo{year}{2019}\natexlab{}.
\newblock \showarticletitle{Variational adversarial active learning}. In
  \bibinfo{booktitle}{\emph{Proceedings of the IEEE/CVF International
  Conference on Computer Vision}}. \bibinfo{pages}{5972--5981}.
\newblock


\bibitem[Springenberg et~al\mbox{.}(2015)]%
        {springenberg2015striving}
\bibfield{author}{\bibinfo{person}{J Springenberg}, \bibinfo{person}{Alexey
  Dosovitskiy}, \bibinfo{person}{Thomas Brox}, {and} \bibinfo{person}{M
  Riedmiller}.} \bibinfo{year}{2015}\natexlab{}.
\newblock \showarticletitle{Striving for Simplicity: The All Convolutional
  Net}. In \bibinfo{booktitle}{\emph{ICLR (workshop track)}}.
\newblock


\bibitem[Stock and Cisse(2018)]%
        {stock2018convnets}
\bibfield{author}{\bibinfo{person}{Pierre Stock} {and}
  \bibinfo{person}{Moustapha Cisse}.} \bibinfo{year}{2018}\natexlab{}.
\newblock \showarticletitle{ConvNets and ImageNet Beyond Accuracy:
  Understanding Mistakes and Uncovering Biases}. In
  \bibinfo{booktitle}{\emph{Proceedings of the European Conference on Computer
  Vision (ECCV)}}. \bibinfo{pages}{498--512}.
\newblock


\bibitem[Tian et~al\mbox{.}(2018)]%
        {tian2018eliminating}
\bibfield{author}{\bibinfo{person}{Maoqing Tian}, \bibinfo{person}{Shuai Yi},
  \bibinfo{person}{Hongsheng Li}, \bibinfo{person}{Shihua Li},
  \bibinfo{person}{Xuesen Zhang}, \bibinfo{person}{Jianping Shi},
  \bibinfo{person}{Junjie Yan}, {and} \bibinfo{person}{Xiaogang Wang}.}
  \bibinfo{year}{2018}\natexlab{}.
\newblock \showarticletitle{Eliminating Background-Bias for Robust Person
  Re-Identification}. In \bibinfo{booktitle}{\emph{Proceedings of the IEEE
  Conference on Computer Vision and Pattern Recognition}}.
  \bibinfo{pages}{5794--5803}.
\newblock


\bibitem[Tommasi and Tuytelaars(2014)]%
        {tommasi2014testbed}
\bibfield{author}{\bibinfo{person}{Tatiana Tommasi} {and}
  \bibinfo{person}{Tinne Tuytelaars}.} \bibinfo{year}{2014}\natexlab{}.
\newblock \showarticletitle{A testbed for cross-dataset analysis}. In
  \bibinfo{booktitle}{\emph{European Conference on Computer Vision}}. Springer,
  \bibinfo{pages}{18--31}.
\newblock


\bibitem[Torralba and Efros(2011)]%
        {torralba2011unbiased}
\bibfield{author}{\bibinfo{person}{Antonio Torralba} {and}
  \bibinfo{person}{Alexei~A Efros}.} \bibinfo{year}{2011}\natexlab{}.
\newblock \showarticletitle{Unbiased look at dataset bias}. In
  \bibinfo{booktitle}{\emph{Computer Vision and Pattern Recognition (CVPR),
  2011 IEEE Conference on}}. IEEE, \bibinfo{pages}{1521--1528}.
\newblock


\bibitem[Wang et~al\mbox{.}(2021)]%
        {wang2021autods}
\bibfield{author}{\bibinfo{person}{Dakuo Wang}, \bibinfo{person}{Josh Andres},
  \bibinfo{person}{Justin~D Weisz}, \bibinfo{person}{Erick Oduor}, {and}
  \bibinfo{person}{Casey Dugan}.} \bibinfo{year}{2021}\natexlab{}.
\newblock \showarticletitle{AutoDS: Towards Human-Centered Automation of Data
  Science}. In \bibinfo{booktitle}{\emph{Proceedings of the 2021 CHI Conference
  on Human Factors in Computing Systems}}. \bibinfo{pages}{1--12}.
\newblock


\bibitem[Wang and Shang(2014)]%
        {6889457}
\bibfield{author}{\bibinfo{person}{Dan Wang} {and} \bibinfo{person}{Yi Shang}.}
  \bibinfo{year}{2014}\natexlab{}.
\newblock \showarticletitle{A new active labeling method for deep learning}. In
  \bibinfo{booktitle}{\emph{2014 International Joint Conference on Neural
  Networks (IJCNN)}}. \bibinfo{pages}{112--119}.
\newblock
\urldef\tempurl%
\url{https://doi.org/10.1109/IJCNN.2014.6889457}
\showDOI{\tempurl}


\bibitem[Xian et~al\mbox{.}(2019)]%
        {8413121}
\bibfield{author}{\bibinfo{person}{Yongqin Xian}, \bibinfo{person}{Christoph~H.
  Lampert}, \bibinfo{person}{Bernt Schiele}, {and} \bibinfo{person}{Zeynep
  Akata}.} \bibinfo{year}{2019}\natexlab{}.
\newblock \showarticletitle{Zero-Shot Learning—A Comprehensive Evaluation of
  the Good, the Bad and the Ugly}.
\newblock \bibinfo{journal}{\emph{IEEE Transactions on Pattern Analysis and
  Machine Intelligence}} \bibinfo{volume}{41}, \bibinfo{number}{9}
  (\bibinfo{year}{2019}), \bibinfo{pages}{2251--2265}.
\newblock
\urldef\tempurl%
\url{https://doi.org/10.1109/TPAMI.2018.2857768}
\showDOI{\tempurl}


\bibitem[Yang et~al\mbox{.}(2019)]%
        {yang2019directing}
\bibfield{author}{\bibinfo{person}{Xi Yang}, \bibinfo{person}{Bojian Wu},
  \bibinfo{person}{Issei Sato}, {and} \bibinfo{person}{Takeo Igarashi}.}
  \bibinfo{year}{2019}\natexlab{}.
\newblock \showarticletitle{Directing DNNs Attention for Facial Attribution
  Classification using Gradient-weighted Class Activation Mapping.}. In
  \bibinfo{booktitle}{\emph{CVPR Workshops}}. \bibinfo{pages}{103--106}.
\newblock


\bibitem[Yoo and Kweon(2019)]%
        {yoo2019learning}
\bibfield{author}{\bibinfo{person}{Donggeun Yoo} {and} \bibinfo{person}{In~So
  Kweon}.} \bibinfo{year}{2019}\natexlab{}.
\newblock \showarticletitle{Learning loss for active learning}. In
  \bibinfo{booktitle}{\emph{Proceedings of the IEEE/CVF Conference on Computer
  Vision and Pattern Recognition}}. \bibinfo{pages}{93--102}.
\newblock


\bibitem[Yosinski et~al\mbox{.}(2015)]%
        {DBLP:journals/corr/YosinskiCNFL15}
\bibfield{author}{\bibinfo{person}{Jason Yosinski}, \bibinfo{person}{Jeff
  Clune}, \bibinfo{person}{Anh~Mai Nguyen}, \bibinfo{person}{Thomas~J. Fuchs},
  {and} \bibinfo{person}{Hod Lipson}.} \bibinfo{year}{2015}\natexlab{}.
\newblock \showarticletitle{Understanding Neural Networks Through Deep
  Visualization}.
\newblock \bibinfo{journal}{\emph{CoRR}}  \bibinfo{volume}{abs/1506.06579}
  (\bibinfo{year}{2015}).
\newblock
\showeprint[arxiv]{1506.06579}
\urldef\tempurl%
\url{http://arxiv.org/abs/1506.06579}
\showURL{%
\tempurl}


\bibitem[You(2020)]%
        {you2020vuejs}
\bibfield{author}{\bibinfo{person}{E You}.} \bibinfo{year}{2020}\natexlab{}.
\newblock \bibinfo{title}{Vuejs framework}.
\newblock \bibinfo{howpublished}{\url{https://vuejs.org}}.
\newblock


\bibitem[Yuan et~al\mbox{.}(2021)]%
        {yuan2021multiple}
\bibfield{author}{\bibinfo{person}{Tianning Yuan}, \bibinfo{person}{Fang Wan},
  \bibinfo{person}{Mengying Fu}, \bibinfo{person}{Jianzhuang Liu},
  \bibinfo{person}{Songcen Xu}, \bibinfo{person}{Xiangyang Ji}, {and}
  \bibinfo{person}{Qixiang Ye}.} \bibinfo{year}{2021}\natexlab{}.
\newblock \showarticletitle{Multiple instance active learning for object
  detection}. In \bibinfo{booktitle}{\emph{Proceedings of the IEEE/CVF
  Conference on Computer Vision and Pattern Recognition}}.
  \bibinfo{pages}{5330--5339}.
\newblock


\bibitem[Zeiler and Fergus(2014)]%
        {zeiler2014visualizing}
\bibfield{author}{\bibinfo{person}{Matthew~D Zeiler} {and} \bibinfo{person}{Rob
  Fergus}.} \bibinfo{year}{2014}\natexlab{}.
\newblock \showarticletitle{Visualizing and understanding convolutional
  networks}. In \bibinfo{booktitle}{\emph{European conference on computer
  vision}}. Springer, \bibinfo{pages}{818--833}.
\newblock


\bibitem[Zhang et~al\mbox{.}(2020)]%
        {zhang2020state}
\bibfield{author}{\bibinfo{person}{Beichen Zhang}, \bibinfo{person}{Liang Li},
  \bibinfo{person}{Shijie Yang}, \bibinfo{person}{Shuhui Wang},
  \bibinfo{person}{Zheng-Jun Zha}, {and} \bibinfo{person}{Qingming Huang}.}
  \bibinfo{year}{2020}\natexlab{}.
\newblock \showarticletitle{State-relabeling adversarial active learning}. In
  \bibinfo{booktitle}{\emph{Proceedings of the IEEE/CVF Conference on Computer
  Vision and Pattern Recognition}}. \bibinfo{pages}{8756--8765}.
\newblock


\bibitem[Zhang et~al\mbox{.}(2018)]%
        {Zhang2018ExaminingCR}
\bibfield{author}{\bibinfo{person}{Quanshi Zhang}, \bibinfo{person}{Wenguan
  Wang}, {and} \bibinfo{person}{Song-Chun Zhu}.}
  \bibinfo{year}{2018}\natexlab{}.
\newblock \showarticletitle{Examining CNN representations with respect to
  Dataset Bias}.
\newblock \bibinfo{journal}{\emph{CoRR}}  \bibinfo{volume}{abs/1710.10577}
  (\bibinfo{year}{2018}).
\newblock


\bibitem[Zhang and Zhu(2018)]%
        {zhang2018visual}
\bibfield{author}{\bibinfo{person}{Quan-shi Zhang} {and}
  \bibinfo{person}{Song-Chun Zhu}.} \bibinfo{year}{2018}\natexlab{}.
\newblock \showarticletitle{Visual interpretability for deep learning: a
  survey}.
\newblock \bibinfo{journal}{\emph{Frontiers of Information Technology \&
  Electronic Engineering}} \bibinfo{volume}{19}, \bibinfo{number}{1}
  (\bibinfo{year}{2018}), \bibinfo{pages}{27--39}.
\newblock


\bibitem[Zhao et~al\mbox{.}(2019)]%
        {zhao2019large}
\bibfield{author}{\bibinfo{person}{Bo Zhao}, \bibinfo{person}{Yanwei Fu},
  \bibinfo{person}{Rui Liang}, \bibinfo{person}{Jiahong Wu},
  \bibinfo{person}{Yonggang Wang}, {and} \bibinfo{person}{Yizhou Wang}.}
  \bibinfo{year}{2019}\natexlab{}.
\newblock \showarticletitle{A large-scale attribute dataset for zero-shot
  learning}. In \bibinfo{booktitle}{\emph{Proceedings of the IEEE/CVF
  Conference on Computer Vision and Pattern Recognition Workshops}}.
  \bibinfo{pages}{0--0}.
\newblock


\bibitem[Zhou et~al\mbox{.}(2016)]%
        {zhou2016learning}
\bibfield{author}{\bibinfo{person}{Bolei Zhou}, \bibinfo{person}{Aditya
  Khosla}, \bibinfo{person}{Agata Lapedriza}, \bibinfo{person}{Aude Oliva},
  {and} \bibinfo{person}{Antonio Torralba}.} \bibinfo{year}{2016}\natexlab{}.
\newblock \showarticletitle{Learning deep features for discriminative
  localization}. In \bibinfo{booktitle}{\emph{Proceedings of the IEEE
  Conference on Computer Vision and Pattern Recognition}}.
  \bibinfo{pages}{2921--2929}.
\newblock


\end{thebibliography}

\end{document}